\def\year{2022}\relax
\documentclass[letterpaper]{article} 
\usepackage{aaai22}  
\usepackage{times}  
\usepackage{helvet}  
\usepackage{courier}  
\usepackage[hyphens]{url}  
\usepackage{graphicx} 
\usepackage{natbib}  
\usepackage{caption} 

\usepackage{epsfig}
\usepackage{amsmath}
\usepackage{amssymb}
\usepackage{amsfonts}
\usepackage{bbm}
\usepackage{multirow}
\usepackage{setspace}
\usepackage{algorithm}
\usepackage{algpseudocode}
\usepackage{amsmath}
\usepackage{float}
\usepackage{epstopdf}
\usepackage{subfigure}
\usepackage{booktabs}

\DeclareCaptionStyle{ruled}{labelfont=normalfont,labelsep=colon,strut=off} 
\usepackage{newfloat}
\usepackage{listings}
\floatstyle{ruled}
\newfloat{listing}{tb}{lst}{}
\floatname{listing}{Listing}
\title{Contrastive Learning from Extremely Augmented Skeleton Sequences for Self-supervised Action Recognition}
\author{
    Tianyu Guo\textsuperscript{\rm 1},
    Hong Liu\textsuperscript{\rm 1\thanks{Corresponding author}},
    Zhan Chen\textsuperscript{\rm 1},
    Mengyuan Liu\textsuperscript{\rm 2},
    Tao Wang\textsuperscript{\rm 1},
    Runwei Ding\textsuperscript{\rm 1}
    \\
}
\affiliations{
    \textsuperscript{\rm 1} Key Laboratory of Machine Perception, Shenzhen Graduate School, Peking University, China\\
    \textsuperscript{\rm 2} The School of Intelligent Systems Engineering, Sun Yat-sen University, China\\
    \{levigty, taowang\}@stu.pku.edu.cn, \{hongliu, zhanchen\_cz, dingrunwei\}@pku.edu.cn, nkliuyifang@gmail.com

}

\usepackage{bibentry}

\begin{document}

\maketitle

\begin{abstract}
In recent years, self-supervised representation learning for skeleton-based action recognition has been developed with the advance of contrastive learning methods. The existing contrastive learning methods use normal augmentations to construct similar positive samples, which limits the ability to explore novel movement patterns. In this paper, to make better use of the movement patterns introduced by extreme augmentations, a Contrastive Learning framework utilizing Abundant Information Mining for self-supervised action Representation (AimCLR) is proposed. First, the extreme augmentations and the Energy-based Attention-guided Drop Module (EADM) are proposed to obtain diverse positive samples, which bring novel movement patterns to improve the universality of the learned representations. Second, since directly using extreme augmentations may not be able to boost the performance due to the drastic changes in original identity, the Dual Distributional Divergence Minimization Loss (D$^3$M Loss) is proposed to minimize the distribution divergence in a more gentle way. Third, the Nearest Neighbors Mining (NNM) is proposed to further expand positive samples to make the abundant information mining process more reasonable. Exhaustive experiments on NTU RGB+D 60, PKU-MMD, NTU RGB+D 120 datasets have verified that our AimCLR can significantly perform favorably against state-of-the-art methods under a variety of evaluation protocols with observed higher quality action representations. Our code is available at \url{https://github.com/Levigty/AimCLR}.
\end{abstract}

\section{Introduction}

On account that action recognition has very broad application in many fields such as human-computer interaction, video content analysis, and smart surveillance, it has always been a popular research topic in the field of computer vision. Due to the development of depth sensors \cite{kinect} and the human pose estimation algorithms \cite{openpose, alphapose}, skeleton-based action recognition has gradually become a significant branch of action recognition.

In the past few years, most of the existing skeleton-based action recognition methods are based on the supervised learning framework. Whether it is a CNN-based method \cite{XYZ1, XYZ2, PR}, RNN-based method \cite{HB-RNN, ST-LSTM, VA-RNN}, or GCN-based method \cite{ST-GCN, 2s-AGCN, AGC-LSTM, MST-GCN}, numerous labeled data is used to learn the action representation. Fully supervised action recognition methods are inevitably data-driven but the cost of labeling large-scale datasets is particularly high. Therefore, more and more researchers intend to use unlabeled skeleton data for learning human action representation.

Recently, several works \cite{LongGAN, PandC, MS2L} focus on designing pretext tasks for self-supervised methods to learn action representations from unlabeled skeleton data. With the development of contrastive self-supervised learning and its ability to make feature representations have better discrimination, several works \cite{AS-CAL, crossclr} directly rely on the contrastive learning framework, using normal augmentations to construct similar positive samples. However, those carefully designed augmentations limit the model to further explore the novel movement patterns exposed by other augmentations and there are still several significant motivations that need to be carefully considered:

\textbf{1) Stronger data augmentations could benefit representation learning.} In SkeletonCLR \cite{crossclr}, it just uses two data augmentations \textit{Shear} and \textit{Crop}. Nevertheless, studies \cite{Infomin, clsa} have shown that data augmentation design is crucial, and the abundant semantic information introduced by stronger data augmentations can significantly improve the generalizability of learned representations and eventually bridge the gap with the fully supervised methods.

\textbf{2) Directly using stronger augmentations could deteriorate the performance.} Stronger data augmentations bring novel movement patterns while the augmented skeleton sequence may not keep the identity of the original sequence. Therefore, directly using extreme augmentations may not necessarily be able to boost the performance due to the drastic changes in original identity. Thus, additional efforts are needed to explore the role of stronger augmentations.

\textbf{3) How to force the model to learn more features.} Simply relying on the contrastive learning framework can not force the model to study more features well. Studies \cite{videomoco, decoupling} have shown that the drop mechanism can be used for contrastive learning, and can effectively solve the problem of over-fitting. Currently, the drop mechanism is not well exploited in self-supervised skeleton-based action recognition.

\textbf{4) How to better expand the positive set to make the learning process more reasonable.} In contrastive learning, two different augmented samples from the same sample are considered as positive samples, while samples in the memory bank are all treated as negative samples. However, the samples in the memory bank are not necessarily all negative samples which makes the learning process unreasonable to a certain extent.

\begin{figure*}[t]

\centerline{\includegraphics[width=17.5cm]{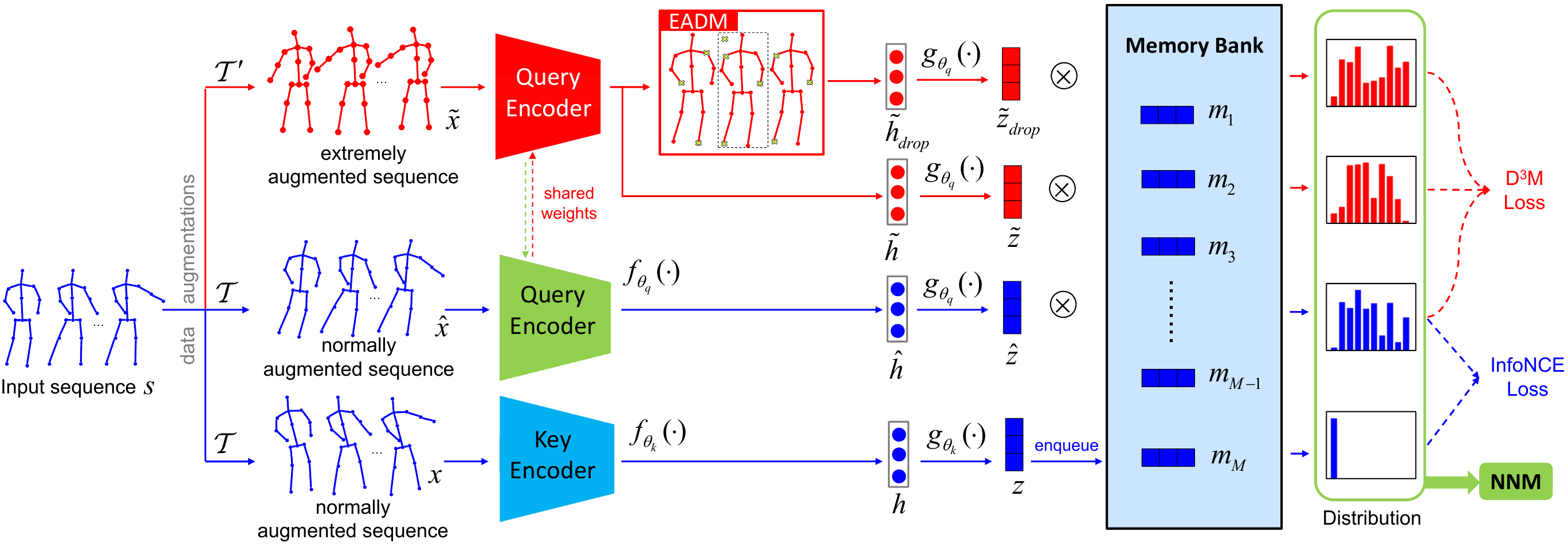}}
\caption{The pipeline of the proposed AimCLR. Through the extreme augmentations $\mathcal{{\mathcal{T}}'}$ and normal augmentations $\mathcal{T}$, $x$, $\hat x$ and $\tilde x$ are obtained from the input sequence $s$. The query encoder and an MLP extract $\hat z$ and $\tilde z$ while the query encoder with EADM and an MLP is used to obtain the $\tilde z_{drop}$.
The momentum updated key encoder and an MLP is used to obtain $z$, $z$ are stored in the memory bank in each training step, serving as negative samples for the next training steps. While using InfoNCE loss, we also propose D$^3$M Loss to minimize the distribution divergence of $\hat z$, $\tilde z$, and $\tilde z_{drop}$. Furthermore, we propose NNM to expand the positive set to make the learning process more reasonable.}
\label{framework}

\end{figure*}

To this end, a contrastive learning framework utilizing abundant information mining for self-supervised action representation (AimCLR) is proposed. Specifically, the framework of AimCLR is shown in Figure \ref{framework}. Different from traditional contrastive learning methods \cite{AS-CAL, crossclr} which directly use normally augmented view, a novel framework proposed in our work is based on the extreme augmentations and the drop mechanism which obtain diverse positive samples and bring abundant spatio-temporal information. Then the Dual Distributional Divergence Minimization Loss (D$^3$M Loss) is proposed to minimize the distribution divergence between the normally augmented view and the extremely augmented views. Furthermore, a Nearest Neighbors Mining (NNM) is used to expand the positive set to make the learning process more reasonable.

In summary, we have made the following contributions:

\begin{itemize}
\item Compared with the traditional contrastive learning method using similar augmented pairs, AimCLR is proposed to use more extreme augmentations and more reasonable abundant information mining which greatly improve the effect of contrastive learning.

\item Specifically, the extreme augmentations and the Energy-based Attention-guided Drop Module (EADM) are proposed to introduce novel movement patterns to force the model to learn more general representations. Then the D$^3$M Loss is proposed to gently learn from the introduced movement patterns. In order to alleviate the irrationality of the positive set, we further propose the Nearest Neighbors Mining (NNM) strategy.

\item With the multi-stream fusion scheme, our 3s-AimCLR achieves state-of-the-art performances under a variety of evaluation protocols such as KNN evaluation, linear evaluation, semi-supervised evaluation, and finetune evaluation protocol on three benchmark datasets.
\end{itemize}

\section{Related Work} \label{section2}

\textbf{Supervised Skeleton-based Action Recognition.} Early skeleton-based action recognition methods are usually based on hand-crafted features \cite{Actionlet, Lie-Group, Rolling}. With the rapid development of deep learning in recent years, some methods \cite{HB-RNN, ST-LSTM, VA-RNN} use RNN to process skeleton data. Meanwhile, several methods convert the 3D skeleton sequence into an image representation and have achieved good results based on CNN \cite{XYZ1, XYZ2, PR}. In recent years, with the introduction of graph convolutional networks, a variety of GCN-based methods \cite{2s-AGCN, AGC-LSTM, MST-GCN} have emerged on the basis of ST-GCN \cite{ST-GCN} to better model the spatio-temporal structure relationship. In this paper, we adopt the widely-used ST-GCN as the encoder to extract the skeleton features.

\textbf{Contrastive Self-Supervised Representation Learning.} Some contrastive learning methods \cite{color, context, rotate} focus on designing various novel pretext tasks to find the pattern information hidden in the unlabeled data. MoCo and MoCov2 \cite{moco, mocov2} promotes contrastive self-supervised learning through a queue-based memory bank and momentum update mechanism. SimCLR \cite{SimCLR} uses a much larger batch size to compute the embeddings in real-time, and uses a multi-layer perceptron (MLP) to further improve the performance of self-supervised representation learning. Current work CLSA \cite{clsa} shows that strong augmentations are beneficial to the performance of downstream tasks and it expects to learn from strongly augmented samples. The development of contrastive self-supervised representation learning also laid the foundation for our AimCLR.

\textbf{Self-supervised Skeleton-based Action Recognition.} LongT GAN \cite{LongGAN} proposes to use the encoder-decoder to regenerate the input sequence to obtain useful feature representation. P\&C \cite{PandC} proposes a training strategy to weaken the decoder, forcing the encoder to learn more discriminative features. \citet{cloud} design a novel skeleton cloud colorization technique to learn skeleton representations. AS-CAL \cite{AS-CAL} and SkeletonCLR \cite{crossclr} use momentum encoder for contrastive learning with single-stream skeleton sequence while CrosSCLR \cite{crossclr} proposes cross-stream knowledge mining strategy to improve the performance and ISC \cite{isc} proposes inter-skeleton contrastive learning to learn from multiple different input skeleton representations. In order to learn more general features, MS$^2$L \cite{MS2L} introduces multiple self-supervised tasks to learn more general representations. However, the currently existing methods rarely explore the gains that abundant spatio-temporal information brings to the task of action recognition. Therefore, a more concise and general framework needs to be proposed.

\section{AimCLR} \label{section3}

\subsection{SkeletonCLR Overview}

SkeletonCLR \cite{crossclr} is based on the recent advanced practice MoCov2 \cite{mocov2} to learn single-stream 3D action representations. The pipeline of the SkeletonCLR is shown in the bottom blue part of Figure \ref{framework}. Given an encoded query $\hat z$ and encoded key $z$, the batch embeddings of $z$ are stored in first-in-first-out memory bank $\textbf{M} = \left\{ {{m_i}} \right\}_{i = 1}^M$ to get rid of redundant computation. It serves as negative samples for the next training steps. Then the InfoNCE loss \cite{cpc} can be written as:

\begin{equation}
\mathcal{L}_{Info} =  - \log \frac{{\exp (\hat z \cdot {z}/\tau )}}{{\exp (\hat z \cdot {z}/\tau )}+{\sum\nolimits_{i = 1}^M {\exp (\hat z \cdot {m_i}/\tau )} }},
\label{eq1}
\end{equation}
where $\tau$ is the temperature hyper-parameter, and dot product $\hat z \cdot z$ is to compute their similarity where $z$, $\hat z$ are normalized.

After computing the InfoNCE loss in Eq. \ref{eq1}, the query encoder is updated via gradients while the key encoder is updated as a moving-average of the query encoder. We denote the parameters of the query encoder as $\theta _q$ and those of the key encoder as $\theta _k$. Then the key encoder is updated as:
\begin{equation}
{\theta _k} \leftarrow m{\theta _k} + (1 - m){\theta _q},
\label{eq2}
\end{equation}
where $m \in \left[ {0,1} \right)$ is a momentum coefficient. The momentum encoder is updated slowly based on the encoder change, which ensures stable key representations.

\subsection{Data Augmentations}

In contrastive learning, augmentations for positive samples bring semantic information for the encoder to learn. However, those carefully designed augmentations limit the encoder to further explore the novel patterns exposed by other augmentations. Therefore, we aim to explore a more general framework in which extreme augmentations can introduce more novel movement patterns than normal augmentations.

\textit{1) Normal Augmentations $\mathcal{T}$.} One spatial augmentation \textit{Shear} and one temporal augmentation \textit{Crop} are used as the normal augmentations like SkeletonCLR.

\textit{2) Extreme Augmentations $\mathcal{{\mathcal{T}}'}$.} We introduce four spatial augmentations: \textit{Shear}, \textit{Spatial Flip}, \textit{Rotate}, \textit{Axis Mask} and two temporal augmentations: \textit{Crop}, \textit{Temporal Flip} and two spatio-temporal augmentations: \textit{Gaussian Noise} and \textit{Gaussian Blur}. We use the combination of all the 8 augmentations (2 normal and 6 other augmentations) as ``Extreme Augmentations'' to finally get one extremely augmented sequence. On account that the combination of extreme augmentations is complicated, we hope to explore a more general framework in which extreme augmentations can introduce more novel movement patterns than normal augmentations.

\subsection{Energy-based Attention-guided Drop Module}

For a feature learned by the encoder, we hope that even if some important features are discarded, different actions can be distinguished. Studies \cite{videomoco, decoupling} have shown that the drop mechanism can be used for contrastive learning, and can effectively solve the problem of over-fitting. It inspires us to calculate the spatio-temporal attention map to drop several important features, which could force the model to learn more features and obtain more general and robust feature representations.

Actually, there are lots of modules \cite{se, cbam, srm} proposed to calculate the attention maps. In order not to introduce additional parameters, we adopt the parameter-free attention module Simam \cite{simam} to calculate the attention map. Formally, $t$ and $x_i$ are the target neuron and other neurons in a single channel of the input feature $X \in {\mathbb{R}^{C \times T \times V}}$ where $C$ denotes the number of channels, $T$ denotes the temporal dimension and $V$ denotes the spatial dimension. The minimal energy $e_t$ of target neuron $t$ can be computed with the following:

\begin{equation}
e_t = \frac{{4({{\hat \sigma }^2} + \lambda )}}{{{{(t - \hat \mu )}^2} + 2{{\hat \sigma }^2} + 2\lambda }},
\label{eq3}
\end{equation}
where $\hat \mu {\rm{ = }}\frac{1}{N}\sum\nolimits_{i = 1}^N {{x_i}}$, ${\hat \sigma ^2}{\rm{ = }}\frac{1}{N}\sum\nolimits_{i = 1}^N {{{({x_i} - \hat \mu )}^2}}$, $\lambda$ is a hyper-parameter, and $N = T \times V$ is the number of neurons on the channel. Eq. \ref{eq3} indicates that the lower energy $e_t$, the neuron $t$ is more distinctive from surround neurons, and more important for visual processing. Therefore, the importance of each neuron can be obtained by $1/e_t$. The we can obtain the attention map $\tilde \alpha$ by that $\tilde \alpha {\rm{ = sigmoid}}(1/\textbf{E})$, where $\textbf{E}$ groups all $e_t$. After that, we use the attention map $\tilde \alpha$ to drop some important features using Algorithm \ref{alg:Drop}.

\begin{algorithm}[t]
  \caption{Energy-based attention-guided drop module.}
  \label{alg:Drop}
  \begin{algorithmic}[1]
    \Require
      a GCN feature $X$: the dimension is ${\mathbb{R}^{C \times T \times V}}$;
      $keep\_margin$: control the importance margin to drop.
    \Ensure
      Features after processing.
    \State Compute the attention map $\tilde \alpha$;
    \State Generate the spatial attention map $M_s$ and the temporal attention map $M_t$ using $\tilde \alpha$ and $keep\_margin$;
    \State Apply the spatial mask: $X{\rm{ = }}X \times M_s$;
    \State Normalize the feature:
    \Statex \qquad $X{\rm{ = }}X \times count({M_s})/count\_ones({M_s})$;
    \State Apply the temporal mask: $X{\rm{ = }}X \times M_t$;
    \State Normalize the feature:
    \Statex \qquad $X{\rm{ = }}X \times count({M_t})/count\_ones({M_t})$;
    \State \Return $X$;
  \end{algorithmic}
\end{algorithm}

\subsection{Dual Distributional Divergence Minimization}

As shown in Figure \ref{framework}, for the input sequence $s$, we apply normal augmentations $\mathcal{T}$ and extreme augmentations $\mathcal{{\mathcal{T}}'}$ to obtain $x$, $\hat x$ and $\tilde x$. The query encoder ${f_{{\theta _q}}}$ is applied to extract features: $\hat h = {f_{{\theta _q}}}(\hat x)$ and $\tilde h = {f_{{\theta _q}}}(\tilde x)$. $\tilde h_{drop}$ is the dropped features after EADM. An MLP head ${g_{{\theta _q}}}$ is applied to project the feature to a lower dimension space: $\hat z = {g_{{\theta _q}}}(\hat h)$, $\tilde z = {g_{{\theta _q}}}(\tilde h)$, ${\tilde z_{drop}} = {g_{{\theta _q}}}({\tilde h_{drop}})$. The key encoder ${f_{{\theta _k}}}$ and ${g_{{\theta _k}}}$ are the momentum updated version of ${f_{{\theta _q}}}$ and ${g_{{\theta _q}}}$.

The memory bank $\textbf{M} = \left\{ {{m_i}} \right\}_{i = 1}^M$ of $M$ negative samples is a first-in-first-out queue updated per iteration by $z$. After each inference step, $z$ will enqueue while the earliest embedding in $\textbf{M}$ will dequeue. $\textbf{M}$ provides numerous negative embeddings while the new calculated $z$ is the positive embedding. Thus, we can obtain a conditional distribution:

\begin{equation}
p({m_i}\left| {{{\hat z}}} \right.) = \frac{{\exp ({{\hat z}} \cdot {m_i}/\tau )}}{{\exp ({{\hat z}} \cdot {z}/\tau ) + \sum\nolimits_{i = 1}^M {\exp ({{\hat z}} \cdot {m_i}/\tau )} }},
\label{eq4}
\end{equation}
which encodes the likelihood of the query ${\hat z}$ being assigned to the embedding $m_i$ in the memory bank $\textbf{M}$. Similarly, we can also have the likelihood of positive pairs for the query ${\hat z}$ being assigned to its positive counterpart $z$:

\begin{equation}
p({z}\left| {{{\hat z}}} \right.) = \frac{{\exp ({{\hat z}} \cdot {z}/\tau )}}{{\exp ({{\hat z}} \cdot {z}/\tau ) + \sum\nolimits_{i = 1}^M {\exp ({{\hat z}} \cdot {m_i}/\tau )} }}.
\label{eq5}
\end{equation}
The InfoNCE loss in Eq. \ref{eq1} can be rewritten in another form:

\begin{equation}
\begin{array}{l}
{{\cal L}_{Info}}{\rm{ = }} - q(z\left| {\hat z} \right.)\log p(z\left| {\hat z} \right.)\\
\;\;\;\;\;\;\;\;\;\;\;\;\;\;\;\;\;\;\;\; - \sum\limits_{i = 1}^M {q({m_i}\left| {\hat z} \right.)\log p({m_i}\left| {\hat z} \right.)},
\end{array}
\label{eq6}
\end{equation}
where $q({z}\left| {{{\hat z}}} \right.)$ is the ideal distribution of the likelihood, $p({z}\left| {{{\hat z}}} \right.)$
is the distribution learned by the network. To avoid the unknown ideal distribution exploration, InfoNCE loss regards $\hat z$ as a one-hot distribution, where positive pairs have $q(z\left| {\hat z} \right.) = 1$ and negative pairs satisfy $q({m_i}\left| {\hat z} \right.) = 0(i \in [1,M])$. It means that InfoNCE loss maximizes the agreement of two different augmented sequences' representations from the same sequence while minimizing the agreement with other negative sequences. To explore the novel movement patterns from the extreme augmentations, a straightforward approach is directly using the extremely augmented sequence as query and the normally augmented sequence as key in InfoNCE loss. However, compared to the normally augmented sequence, the extremely augmented sequence may not keep the identity of the original sequence due to the dramatic changes in movement patterns, leading to performance degradation.

In addition, it's almost impossible to obtain the ideal likelihood distribution. Fortunately, CLSA \cite{clsa} found that the normally augmented query and the extremely augmented query share similar distribution for a randomly initialized network. It inspires us that the distribution of normally augmented query over memory bank can be used to supervise that of the extremely augmented query. It avoids directly using one-hot distribution for extremely augmented views and is able to explore the novel patterns exposed by the extreme augmentations.

Similar to Eq. \ref{eq4} and Eq. \ref{eq5}, we obtain a conditional distribution for $\tilde z$ based on its positive samples and negative samples: $p({z}\left| {{{\tilde z}}} \right.)$ and $p({m_i}\left| {{{\tilde z}}} \right.)$. The conditional distribution ${p({m_i}\left| {\tilde z_{drop}} \right.)}$ and $p({z}\left| {\tilde z_{drop}} \right.)$ for $\tilde z_{drop}$ is calculated in the same way. Then, we propose to minimize the following distributional divergence between the normally augmented view and the extremely augmented view such that:

\begin{equation}
\begin{array}{l}
{{\cal L}_{d1}}{\rm{ = }} - p(z\left| {\hat z} \right.)\log p(z\left| {\tilde z} \right.)\\
\;\;\;\;\;\;\;\;\;\;\;\;\;\;\;\;\;\;\; - \sum\limits_{i = 1}^M {p({m_i}\left| {\hat z} \right.)\log p({m_i}\left| {\tilde z} \right.)}.
\end{array}
\label{eq7}
\end{equation}
Similarly, the distributional divergence between the normally augmented view and the dropped extremely augmented view is minimized such that:

\begin{equation}
\begin{array}{l}
{{\cal L}_{d2}}{\rm{ = }} - p(z\left| {\hat z} \right.)\log p(z\left| {{{\tilde z}_{drop}}} \right.)\\
\;\;\;\;\;\;\;\;\;\;\;\;\;\;\;\;\;\;\; - \sum\limits_{i = 1}^M {p({m_i}\left| {\hat z} \right.)\log p({m_i}\left| {{{\tilde z}_{drop}}} \right.)}.
\end{array}
\label{eq8}
\end{equation}
Therefore, the proposed D$^3$M loss can be formulated as ${{\cal L}_D}{\rm{ = }}1/2{\rm{(}}{{\cal L}_{d1}}{\rm{ + }}{{\cal L}_{d2}})$.

\subsection{Nearest Neighbors Mining}\label{section3.5}

Traditional contrastive learning methods regard the normally augmented samples from the same sample as positive samples and all samples in the memory bank as negative samples. However, the samples in the memory bank are not necessarily all negative samples \cite{nnclr}. Therefore, we hope that the nearest neighbors of query $\hat z$, $\tilde z$, and $\tilde z_{drop}$ over the memory bank $\textbf{M}$ should be considered as positive samples to expand the positive set.

Specifically, $N_ + ^n$ is the index set of the nearest top-k neighbors that are most similar to the normally augmented query $\hat z$ in the memory bank $\textbf{M}$. Similarly, we could also have the $N_ + ^e$ and $N_ + ^d$ to represents the index set of the nearest top-k neighbors of the extremely augmented query $\tilde z$ and the dropped extremely augmented query $\tilde z_{drop}$. Thus, we can set the nearest top-k neighbors as positive samples to make the learning process more reasonable:

\begin{equation}
{{\cal L}_{N}} =  - \log \frac{{\exp (\hat z \cdot z/\tau ){\rm{ + }}\sum\nolimits_{i \in {N_ + }} {\exp (\hat z \cdot {m_i}/\tau )} }}{{\exp (\hat z \cdot z/\tau ) + \sum\nolimits_{i = 1}^M {\exp (\hat z \cdot {m_i}/\tau )} }},
\label{eq9}
\end{equation}
where ${N_ + } = N_ + ^n \cup N_ + ^e \cup N_ + ^d$. Compared to Eq. \ref{eq1}, Eq. \ref{eq9} will lead to a more regular space by pulling close more high-confidence positive samples.

\textbf{Two-stage Training Strategy.} In the early training stage, the model is not stable and strong enough to provide reliable nearest neighbors. Thus, we perform two-stage training for AimCLR: In the first stage, the model is trained with the loss function: ${\cal L}_{1}{\rm{ = }}\alpha {{\cal L}_{Info}}{\rm{ + }}\beta {{\cal L}_D}$. Then in the second training stage, the loss function is ${\cal L}_{2}{\rm{ = }}\alpha {{\cal L}_{N}}{\rm{ + }}\beta {{\cal L}_D}$ to start mining the nearest neighbors. Here, $\alpha$ and $\beta$ are the coefficient to balance the loss. Though other values may achieve better results, we use $\alpha = \beta = 1$ to make AimCLR more general.

\section{Experiments}\label{section4}

\subsection{Dataset}

\textbf{PKU-MMD Dataset} \cite{pkummd}: It contains almost 20,000 action sequences covering 51 action classes. It consists of two subsets. Part \uppercase\expandafter{\romannumeral1} is an easier version for action recognition, while part \uppercase\expandafter{\romannumeral2} is more challenging with more noise caused by view variation. We conduct experiments under the cross-subject protocol on the two subsets.

\textbf{NTU RGB+D 60 Dataset} \cite{ntu60}: The dataset contains 56,578 action sequences and 60 action classes. There are two evaluation protocols: cross-subject (xsub) and cross-view (xview). In xsub, half of the subjects are used as training sets, and the rest are used as test sets. In xview, the samples of camera 2 and 3 are used for training while the samples of camera 1 are used for testing.

\textbf{NTU RGB+D 120 Dataset} \cite{ntu120}: It is NTU RGB+D 60 based extension, whose scale is up to 120 action classes and 113,945 sequences. There are two evaluation protocols: cross-subject (xsub) and cross-setup (xset). In xsub, actions performed by 53 subjects are for training and the others are for testing. In xset, all 32 setups are separated as half for training and the other half for testing.

\begin{table}[t]
\centering
\small
\tabcolsep1.5mm
\begin{tabular}{cccc|cc}
\toprule
\multirow{2}{*}{w/ NA} & \multirow{2}{*}{w/ EA} & \multirow{2}{*}{w/ EADM} & \multirow{2}{*}{w/ NNM} & \multicolumn{2}{c}{NTU-60(\%)} \\
                    &                  &             &             & xsub         & xview          \\ \midrule
\checkmark          &                  &             &             & 75.0         & 79.8            \\
                    & \checkmark       &             &             & 71.3         & 77.8            \\
\checkmark          & \checkmark       &             &             & 77.4         & 82.5            \\
\checkmark          & \checkmark       & \checkmark  &             & 78.2         & 82.8             \\
\checkmark          & \checkmark       & \checkmark  & \checkmark  & \textbf{78.9}    & \textbf{83.8}            \\ \bottomrule
\end{tabular}
\caption{Ablation study results on NTU-60 dataset.}
\label{ab}
\end{table}

\begin{table*}
\centering
\small
\tabcolsep2mm
\begin{tabular}{l|l|cc|cc|cc|cc|cc}
\toprule
\multirow{3}{*}{Method} & \multirow{3}{*}{Stream} & \multicolumn{4}{c|}{NTU-60(\%)} & \multicolumn{2}{c|}{PKU(\%)} & \multicolumn{4}{c}{NTU-120(\%)}                         \\ \cline{3-12}
   &     & \multicolumn{2}{c|}{xsub} & \multicolumn{2}{c|}{xview} & \multicolumn{2}{c|}{part I}  & \multicolumn{2}{c|}{xsub} & \multicolumn{2}{c}{xset} \\
   &     & acc.    & $\Delta$  & acc.    & $\Delta$   & acc.     & $\Delta$     & acc.     & $\Delta$  & acc.     & $\Delta$       \\ \midrule
SkeletonCLR (CVPR $21$)    & joint              & 68.3   &     & 76.4    &        & 80.9   &    & 56.8   &     & 55.9   & \\
\textbf{AimCLR (ours)}     & joint    & \textbf{74.3}   & $\uparrow$ 6.0   & \textbf{79.7}    & $\uparrow$ 3.3   & \textbf{83.4}   & $\uparrow$ 2.5  & \textbf{63.4}   & $\uparrow$ 6.6 &   \textbf{63.4}   & $\uparrow$ 7.5 \\ \midrule
SkeletonCLR (CVPR $21$)    & motion             & 53.3   &     & 50.8    &        & 63.4   &    & 39.6   &   & 40.2   &   \\
\textbf{AimCLR (ours)}     & motion   & \textbf{66.8}   & $\uparrow$ 13.5   & \textbf{70.6}    & $\uparrow$ 19.8  & \textbf{72.0}   & $\uparrow$ 8.6  & \textbf{57.3}   & $\uparrow$ 17.7  & \textbf{54.4}   & $\uparrow$ 14.2 \\ \midrule
SkeletonCLR (CVPR $21$)    & bone               & 69.4   &     & 67.4    &        & 72.6   &   & 48.4   &  & 52.0   &     \\
\textbf{AimCLR (ours)}     & bone     & \textbf{73.2}   & $\uparrow$ 3.8   & \textbf{77.0}    & $\uparrow$ 9.6   & \textbf{82.0}   & $\uparrow$ 9.4  & \textbf{62.9}   & $\uparrow$ 14.5 & \textbf{63.4}   & $\uparrow$ 11.4    \\ \midrule
3s-SkeletonCLR (CVPR $21$) & joint+motion+bone  & 75.0   &     & 79.8    &        & 85.3   &   & 60.7   & & 62.6   &      \\
\textbf{3s-AimCLR (ours)}  & joint+motion+bone   & \textbf{78.9}   & $\uparrow$ 3.9   & \textbf{83.8}    & $\uparrow$ 4.0   & \textbf{87.8}   & $\uparrow$ 2.5  & \textbf{68.2}   & $\uparrow$ 7.5 & \textbf{68.8}   & $\uparrow$ 6.2    \\ \bottomrule
\end{tabular}
\caption{Linear evaluation results compared with SkeletonCLR on NTU-60, PKU-MMD, and NTU-120 dataset. ``$\Delta$'' represents the gain compared to SkeletonCLR using the same stream data. ``3s'' means three stream fusion.}
\label{com}
\end{table*}

\subsection{Experimental Settings}

All the experiments are conducted on the PyTorch \cite{pytorch} framework. For data pre-processing, we follow SkeletonCLR and CrosSCLR \cite{crossclr} for a fair comparison. The mini-batch size is set to 128.

\textbf{Self-supervised Pretext training.} ST-GCN is adopted as the encoder. For contrastive settings, we follow that in SkeletonCLR. For optimization, we use SGD with momentum (0.9) and weight decay (0.0001). The model is trained for 300 epochs with a learning rate of 0.1 (decreases to 0.01 at epoch 250). For the nearest neighbors mining, we set $k = 1$. For the two-stage training strategy mentioned in Section \ref{section3.5}, the encoder is trained with ${\cal L}_{1}$ in the first 150 epochs while trained with ${\cal L}_{2}$ in the remaining epochs. For a fair comparison, we also generate three streams of skeleton sequences, i.e., joint, bone, and motion. For all the reported results of three streams, we use the weights of $[0.6, 0.6, 0.4]$ for weighted fusion like other multi-stream GCN methods.

\textbf{KNN Evaluation Protocol.} A K-Nearest Neighbor (KNN) classifier is used on the features of the trained encoder. It can also reflect the quality of the features learned by the encoder.

\textbf{Linear Evaluation Protocol.} The models are verified by linear evaluation for the action recognition task. Specifically, we train a linear classifier (a fully-connected layer followed by a softmax layer) supervised with encoder fixed.

\textbf{Semi-supervised Evaluation Protocol.} We pre-train the encoder with all data and then finetuning the whole model with only 1\% or 10\% randomly selected labeled data.

\textbf{Finetune Protocol.} We append a linear classifier to the trained encoder and then train the whole model to compare it with fully supervised methods.


\begin{figure}[t]
\centering
\subfigure[SkeletonCLR]{\includegraphics[width=2.6cm]{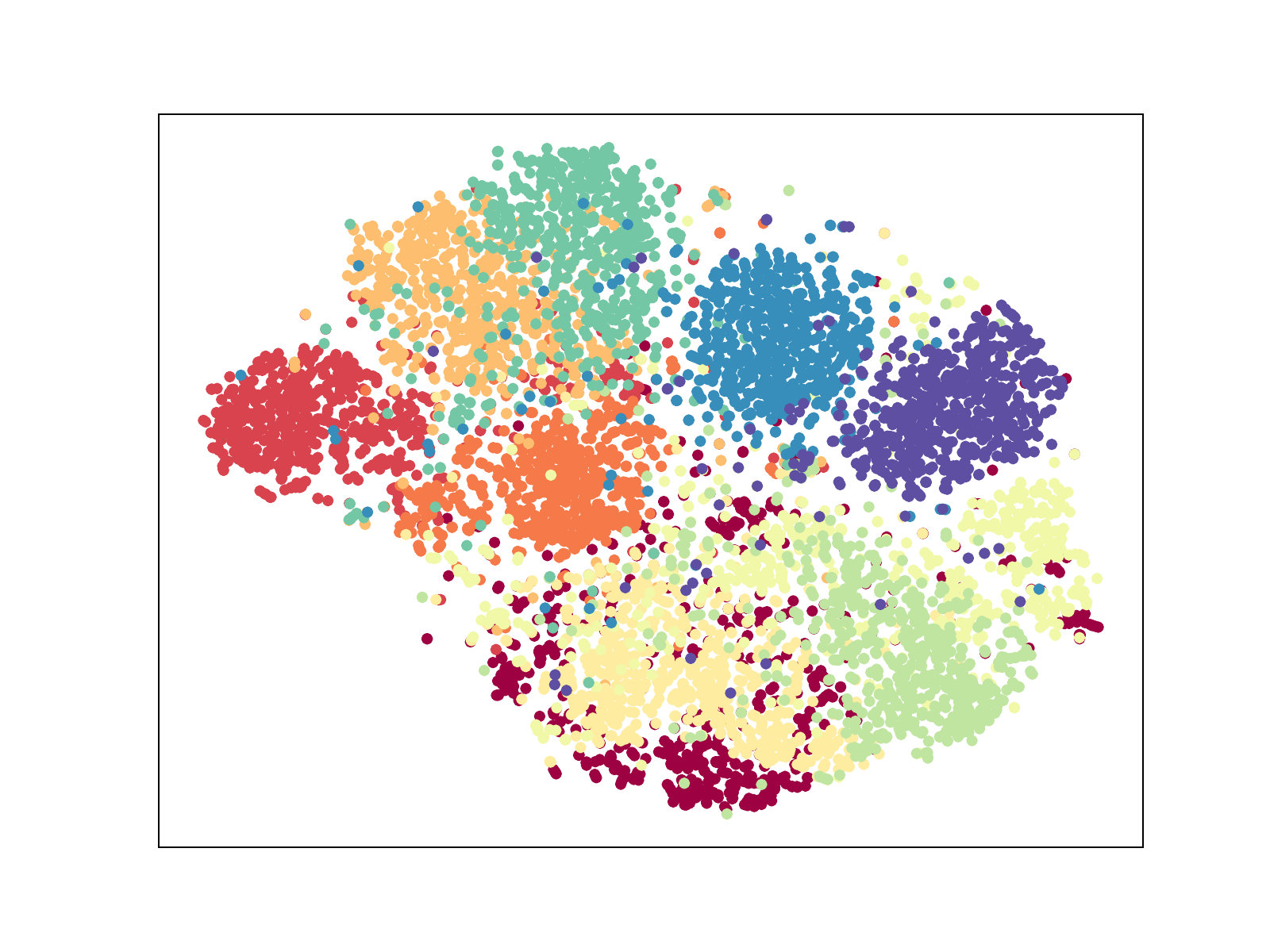}}
\subfigure[3s-SkeletonCLR]{\includegraphics[width=2.6cm]{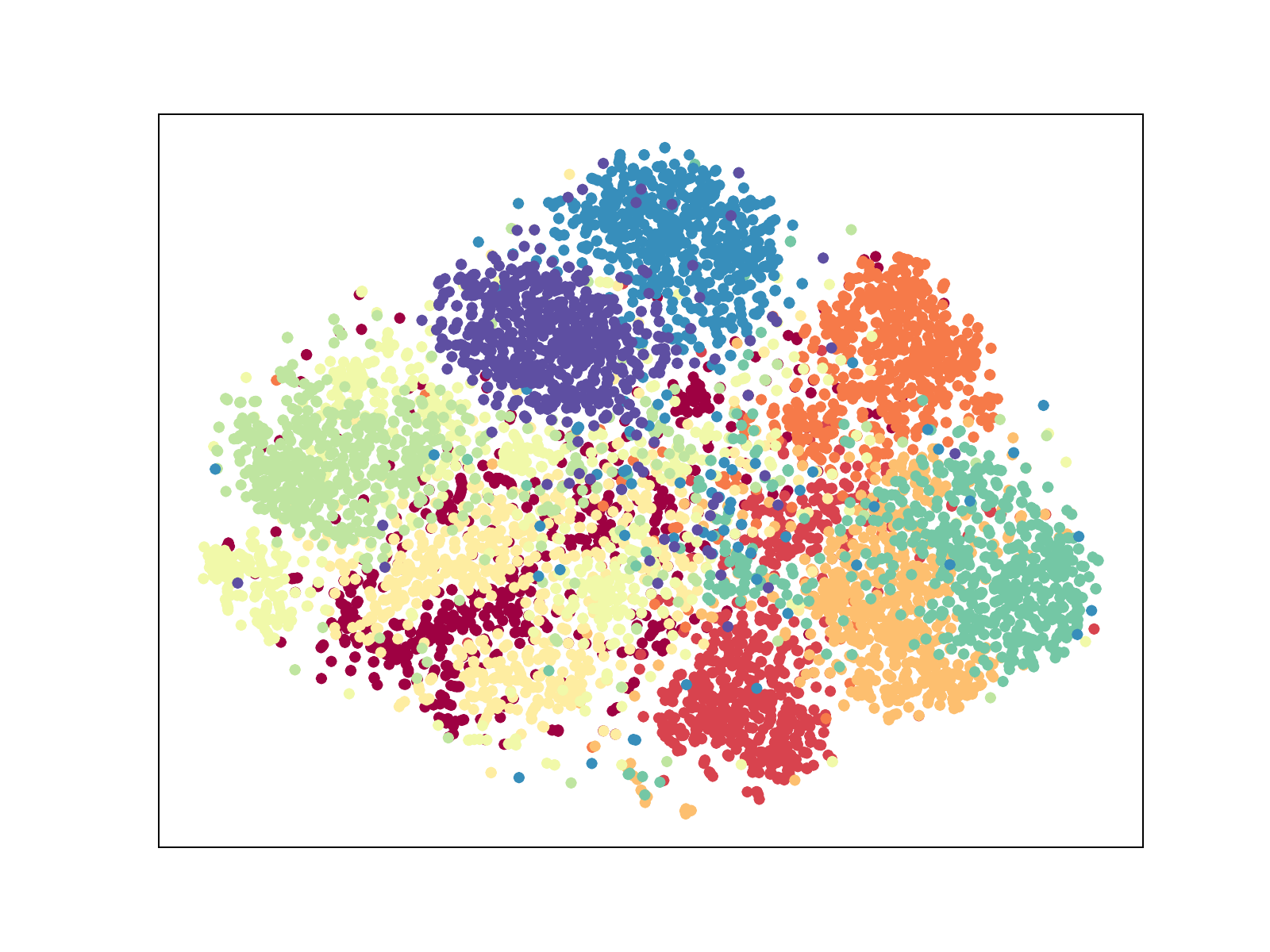}}
\subfigure[3s-CrosSCLR$^\S$]{\includegraphics[width=2.6cm]{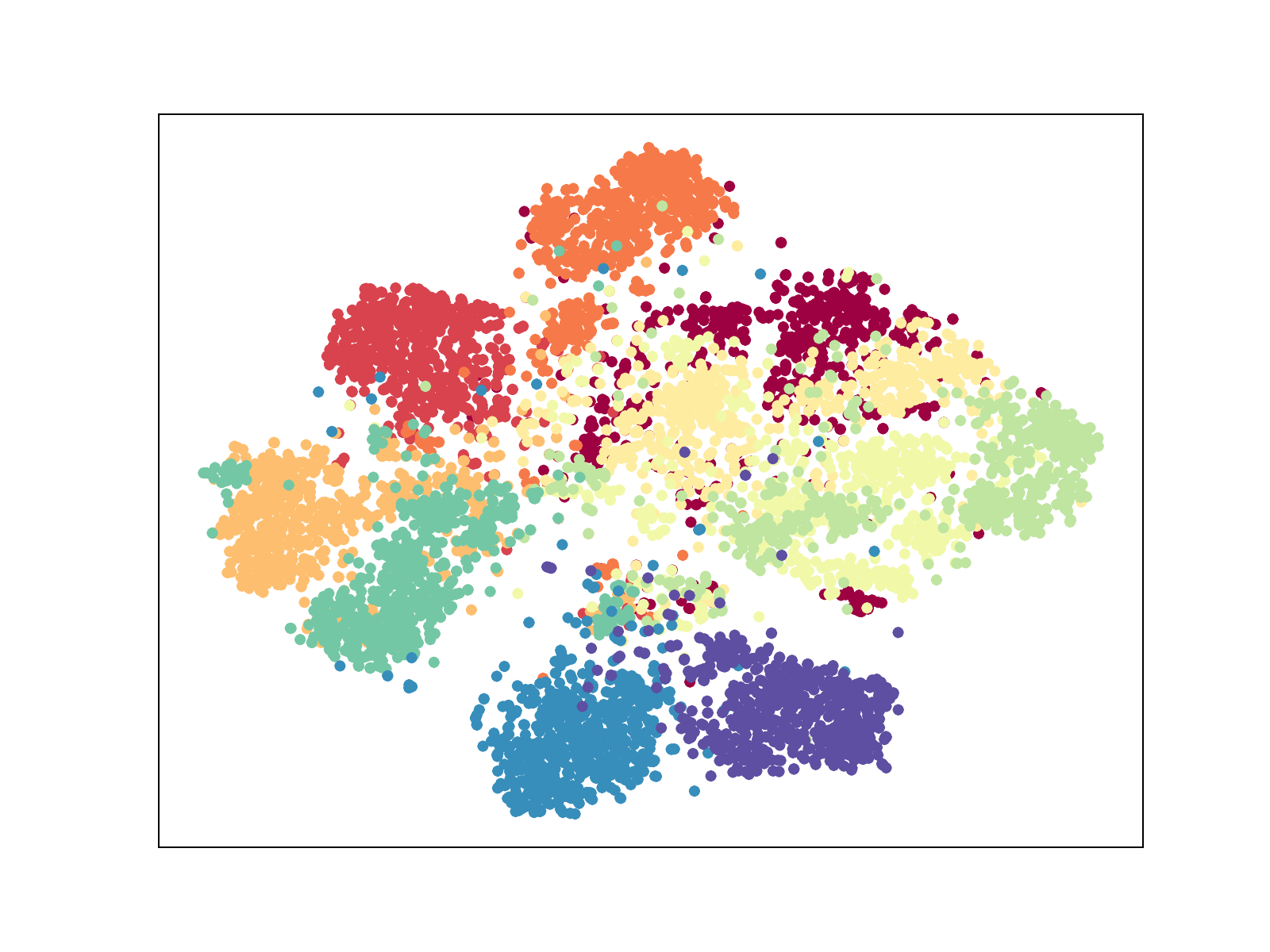}}
\\
\centering
\subfigure[AimCLR]{\includegraphics[width=2.6cm]{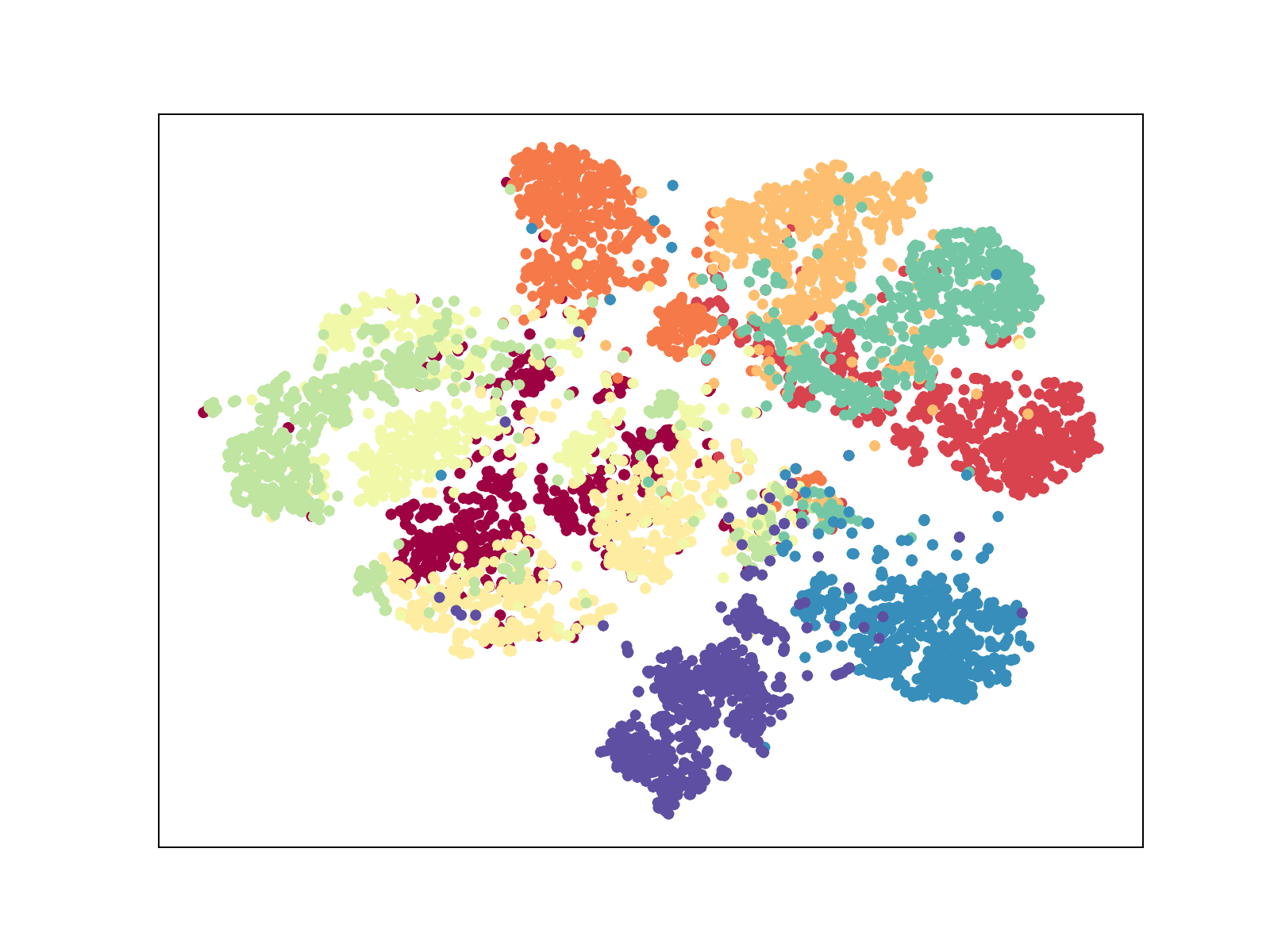}}
\subfigure[3s-AimCLR]{\includegraphics[width=2.6cm]{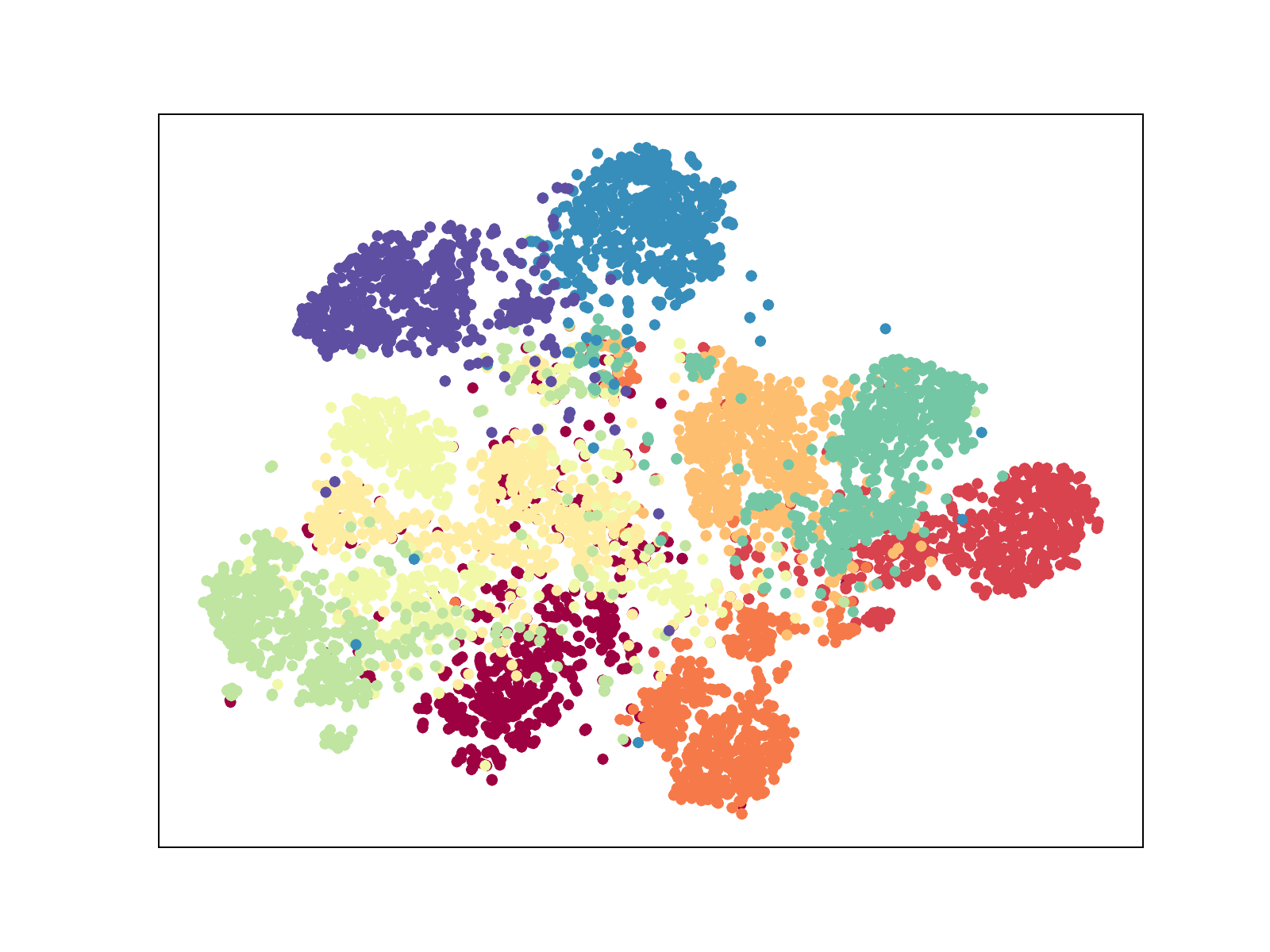}}
\subfigure[3s-AimCLR$^\S$]{\includegraphics[width=2.6cm]{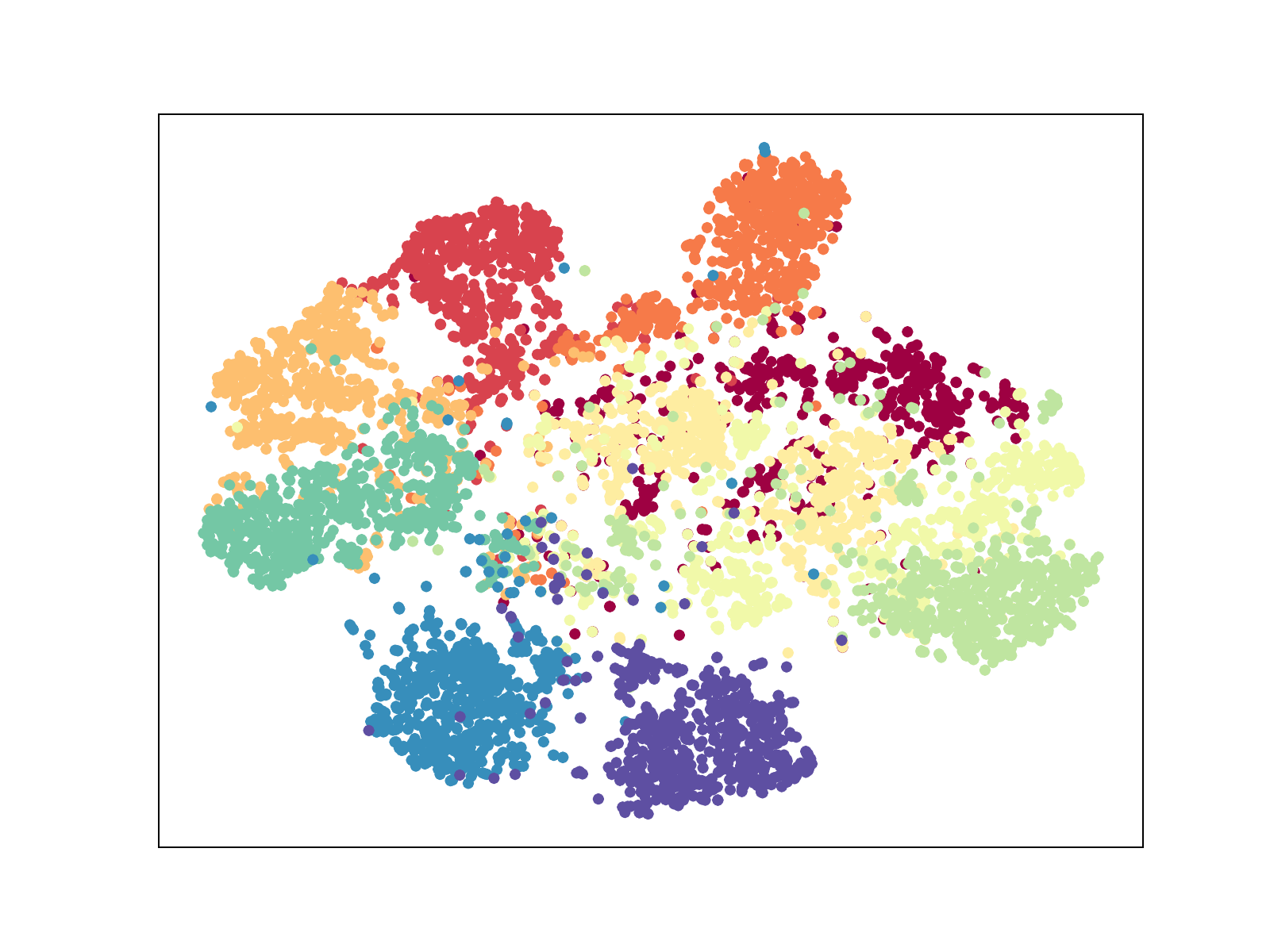}}
\caption{The t-SNE visualization of embeddings on NTU-60 xsub. ``$\S$'' means using cross-stream knowledge mining strategy proposed in 3s-CrosSCLR \cite{crossclr}.}
\label{tsne}
\end{figure}

\begin{table}[t]
\centering
\small
\begin{tabular}{l|cccc}
\toprule
Method                          & 100ep   & 150ep    & 200ep    & 300ep \\ \midrule
3s-SkeletonCLR(repro.)          & 71.3    & 73.8     & 74.1     & 74.1     \\
3s-CrosSCLR(repro.)$^\S$        & 70.0    & 72.8     & 76.0     & 77.2     \\
\textbf{3s-AimCLR (ours)$^\S$}  & 75.4    & 76.0     & 78.2     & 78.6     \\
\textbf{3s-AimCLR (ours)}       & \textbf{76.5} & \textbf{77.4} & \textbf{78.3} & \textbf{78.9}\\ \bottomrule
\end{tabular}
\caption{Linear evaluation results on NTU-60 xsub for different epochs.}
\label{epoch}
\end{table}

\subsection{Ablation Study}

\begin{figure}[t]
\centerline{\includegraphics[width=7.5cm]{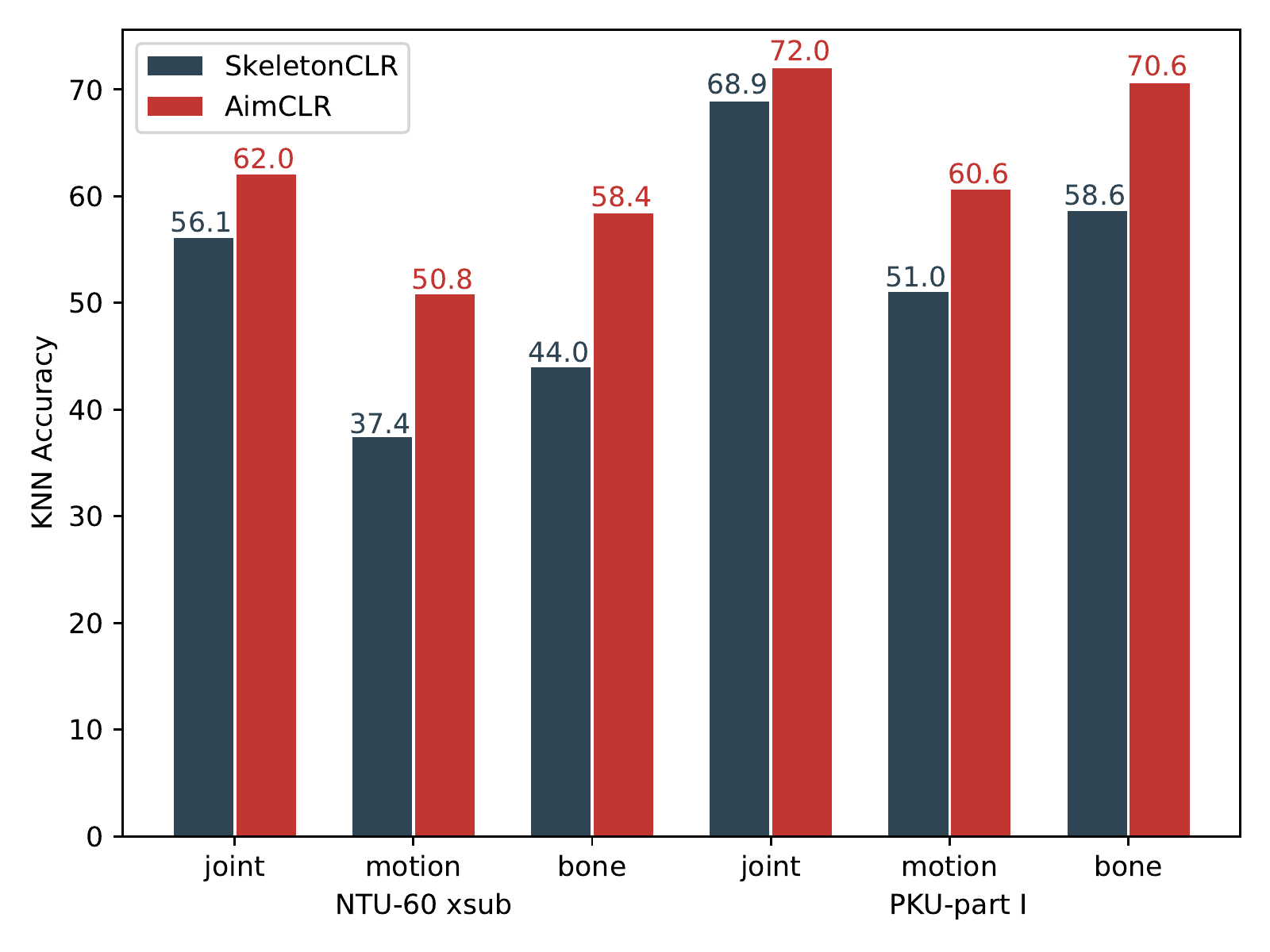}}
\caption{Comparison of KNN accuracy of SkeletonCLR \cite{crossclr} and our AimCLR.}
\label{knn}
\end{figure}

We conduct ablation studies on different datasets to verify the effectiveness of different components of our method.

\textbf{The effectiveness of the data augmentation.} As shown in Table \ref{ab}, 3s-SkeletonCLR \cite{crossclr} uses the normal augmentations (w/ NA) and achieves the accuracy of 75.0\% and 79.8\% on xsub and xview respectively. While simply replacing the normal augmentations with extreme augmentations (w/ EA), the accuracies drop on both xsub and xview. It also illustrates that directly using extreme augmentations may not necessarily be able to boost the performance due to the drastic changes in original identity. While both EA and NA are used, i.e., when ${\cal L}_{d1}$ loss in Eq. \ref{eq7} comes into play, the accuracy is improved by 2.4\% and 2.7\%.

\textbf{The effectiveness of the EADM and NNM.} From Table \ref{ab}, it is worth noting that when EADM is introduced, the accuracy on xsub and xview are improved by 0.8\% and 0.3\%, respectively. Notably, our 3s-AimCLR achieves the highest accuracy when NNM is further introduced. It also shows that the proposed EADM and NNM can make the encoder further learn more robust and suitable features for downstream tasks.

\begin{table}[t]
\centering
\small
\begin{tabular}{l|cc}
\toprule
\multirow{2}{*}{Method}   & \multicolumn{2}{c}{NTU-60(\%)} \\
                                    & xsub           & xview           \\ \midrule
\emph{Single-stream:}               &                &                 \\
LongT GAN (AAAI $18$)             & 39.1           & 48.1            \\
MS$^2$L (ACM MM $20$)              & 52.6           & -               \\
AS-CAL (Information Sciences $21$)& 58.5           & 64.8            \\
P\&C (CVPR $20$)                  & 50.7           & 76.3            \\
SeBiReNet (ECCV $20$)             & -              & \textbf{79.7}   \\
SkeletonCLR (CVPR $21$)           & 68.3           & 76.4            \\
\textbf{AimCLR (ours)}            & \textbf{74.3}  & \textbf{79.7}   \\ \midrule
\emph{Three-stream:}              &                &                 \\
3s-SkeletonCLR (CVPR $21$)        & 75.0           & 79.8              \\
3s-Colorization (ICCV $21$)       & 75.2           & 83.1              \\
3s-CrosSCLR (CVPR $21$)$^\S$      & 77.8           & 83.4              \\
\textbf{3s-AimCLR (ours)$^\S$}    & 78.6           & 82.6              \\
\textbf{3s-AimCLR (ours)}         & \textbf{78.9}  & \textbf{83.8}     \\ \bottomrule
\end{tabular}
\caption{Linear evaluation results on NTU-60 dataset.}
\label{ntu}
\end{table}

\textbf{The effectiveness of the AimCLR.} We conduct experiments on three datasets to verify the performance of our AimCLR and the SkeletonCLR. As can be seen from Table \ref{com}, for the three different streams of the three datasets, AimCLR performs much better than SkeletonCLR. The gains in motion and bone stream are significant. For the fusion results, our 3s-AimCLR far exceeds 3s-SkeletonCLR on the three datasets. In addition, it can be seen from Table \ref{epoch} that our 3s-AimCLR is always better than 3s-CrosSCLR and 3s-SkeletonCLR under the same training epochs no matter using cross-stream knowledge mining strategy or not. The result of 3s-AimCLR at 100 epochs is even better than the result of 3s-SkeletonCLR at 300 epochs.

\textbf{Qualitative Results.} We apply t-SNE \cite{tsne} with fix settings to show the embedding distribution in Figure \ref{tsne}. From the visual results, we can draw a conclusion that 3s-AimCLR could cluster the embeddings of the same class closer than 3s-SkeletonCLR for the simply fusion results. While using the cross-stream knowledge mining strategy, our 3s-AimCLR can also make the action classes that overlapped seriously more distinguishable compared with 3s-CrosSCLR.

\begin{table}[t]
\centering
\small
\begin{tabular}{l|cc}
\toprule
Method & part \uppercase\expandafter{\romannumeral1}(\%) & part \uppercase\expandafter{\romannumeral2}(\%) \\ \midrule
\emph{Supervised:}            &                &                 \\
ST-GCN (AAAI $18$)            & 84.1           & 48.2            \\
VA-LSTM (TPAMI $19$)          & 84.1           & 50.0            \\ \midrule
\emph{Self-supervised:}       &                &                 \\
LongT GAN (AAAI $18$)         & 67.7           & 26.0            \\
MS$^2$L (ACM MM $20$)          & 64.9           & 27.6            \\
3s-CrosSCLR (CVPR $21$)$^\S$  & 84.9           & 21.2            \\
ISC (ACM MM $21$)              & 80.9           & 36.0            \\
\textbf{3s-AimCLR (ours)$^\S$}& 87.4           & \textbf{39.5}   \\
\textbf{3s-AimCLR (ours)}     & \textbf{87.8}  & 38.5            \\ \bottomrule
\end{tabular}
\caption{Linear evaluation results on PKU-MMD dataset.}
\label{pku-com}
\end{table}

\begin{table}[t]
\centering
\small
\begin{tabular}{l|cc}
\toprule
Method                            & xsub(\%)       & xset(\%)        \\      \midrule
P\&C (CVPR $20$)                  & 42.7           & 41.7            \\
AS-CAL (Information Sciences $21$)& 48.6           & 49.2            \\
3s-CrosSCLR (CVPR $21$)$^\S$      & 67.9           & 66.7            \\
ISC (ACM MM $21$)                  & 67.9           & 67.1            \\
\textbf{3s-AimCLR (ours)$^\S$}    & 68.0           & 68.7             \\
\textbf{3s-AimCLR (ours)}         & \textbf{68.2}  & \textbf{68.8}   \\ \bottomrule
\end{tabular}
\caption{Linear evaluation results on NTU-120 dataset.}
\label{ntu120-com}
\end{table}

\subsection{Comparison with State-of-the-art}

We compare the proposed method with prior related methods under a variety of evaluation protocols.

\textbf{KNN Evaluation Results.} Notably, the KNN classifier does not require learning extra weights compared with the linear classifier. From Figure \ref{knn}, we can see that our AimCLR is better than SkeletonCLR on the two datasets under the KNN classifier. The obvious gains also show that the features learned by our AimCLR are more discriminative.

\textbf{Linear Evaluation Results on NTU-60.} As shown in Table \ref{ntu}, for a single stream (i.e., joint stream), our AimCLR outperforms all other methods \cite{LongGAN, MS2L, AS-CAL, PandC, sebirenet, crossclr}. For the performance of the 3-streams, our 3s-AimCLR leads 3s-SkeletonCLR 3.9\% and 4.0\% under the xsub and xview protocols, respectively. It is worth mentioning that regardless of whether 3s-AimCLR uses a cross-stream knowledge mining strategy, the results are better than the 3s-CrossSCLR and 3s-Colorization \cite{cloud}. It also indicates that even without the knowledge mining between streams, 3s-AimCLR has the ability to learn better feature representations.

\textbf{Linear Evaluation Results on PKU-MMD.} As shown in Table \ref{pku-com}, our 3s-AimCLR is ahead of the existing self-supervised methods in both part \uppercase\expandafter{\romannumeral1} and part \uppercase\expandafter{\romannumeral2} of this dataset. Part \uppercase\expandafter{\romannumeral2} is more challenging with more skeleton noise caused by the view variation. Notably, 3s-CrosSCLR suffers on part \uppercase\expandafter{\romannumeral2} while our 3s-AimCLR performs well. It also proves that our 3s-AimCLR has a strong ability to cope with movement patterns caused by skeleton noise.

\textbf{Linear Evaluation Results on NTU-120.} As shown in Table \ref{ntu120-com}, our 3s-AimCLR defeats the other self-supervised methods on NTU-120. Our fusion results outperforms the advanced ISC (68.2\% vs 67.9\% on xsub and 68.8\% vs 67.1\% on xset). This shows that our 3s-AimCLR is also competitive on multi-class large-scale datasets.

\textbf{Semi-supervised Evaluation Results.} From Table \ref{semi}, even only with a small labeled subset, our 3s-AimCLR performs better than the state-of-the-art consistently for all configurations. The results of using 1\% and 10\% labeled data far exceed ISC, 3s-CrosSCLR, and 3s-Colorization. It also proves that the novel movement patterns brought by extreme augmentations have a huge impact when there is only a small amount of labeled data.

\begin{table}[t]
\centering
\small
\tabcolsep1.4mm
\begin{tabular}{l|cc|cc}
\toprule
\multirow{2}{*}{Method}        & \multicolumn{2}{c|}{PKU-MMD(\%)}     & \multicolumn{2}{c}{NTU-60(\%)} \\
& part \uppercase\expandafter{\romannumeral1} & part \uppercase\expandafter{\romannumeral2}  & xsub  & xview  \\ \midrule
\emph{1\% labeled data:}       &               &                &               &               \\
LongT GAN (AAAI $18$)        & 35.8          & 12.4           & 35.2          & -             \\
MS$^2$L (ACM MM $20$)         & 36.4          & 13.0           & 33.1          & -             \\
ISC (ACM MM $21$)             & 37.7          & -              & 35.7          & 38.1             \\
3s-CrosSCLR (CVPR $21$)      & 49.7          & 10.2           & 51.1          & 50.0          \\
3s-Colorization (ICCV $21$)  & -             & -              & 48.3          & 52.5          \\
\textbf{3s-AimCLR (ours)}      & \textbf{57.5}    & \textbf{15.1} & \textbf{54.8} & \textbf{54.3} \\ \midrule
\emph{10\% labeled data:}      &               &                &               &               \\
LongT GAN (AAAI $18$)        & 69.5          & 25.7           & 62.0          & -             \\
MS$^2$L (ACM MM $20$)         & 70.3          & 26.1           & 65.2          & -             \\
ISC (ACM MM $21$)             & 72.1          & -              & 65.9          & 72.5             \\
3s-CrosSCLR (CVPR $21$)      & 82.9          & 28.6           & 74.4          & 77.8           \\
3s-Colorization (ICCV $21$)  & -             & -              & 71.7          & 78.9          \\
\textbf{3s-AimCLR (ours)}      & \textbf{86.1}    & \textbf{33.4}     & \textbf{78.2} & \textbf{81.6}  \\ \bottomrule
\end{tabular}
\caption{Semi-supervised evaluation results on PKU-MMD dataset and NTU-60 dataset.}
\label{semi}
\end{table}

\textbf{Finetuned Evaluation Results.} For fair comparisons, the ST-GCN used in the methods of Table \ref{finetune} all have the same structure and parameters. For a single bone stream, the finetuned results of our AimCLR are better than that of SkeletonCLR. What's more, the finetuned 3s-AimCLR also outperforms the 3s-CrosSCLR and the supervised 3s-ST-GCN, indicating the effectiveness of our method.

\begin{table}[t]
\centering
\small
\begin{tabular}{l|cccc}
\toprule
\multirow{2}{*}{Method} & \multicolumn{2}{c}{NTU-60(\%)} & \multicolumn{2}{c}{NTU-120(\%)} \\
                                    & xsub           & xview           & xsub            & xset            \\ \midrule
SkeletonCLR (CVPR $21$)$^\dag$     & 82.2           & 88.9            & 73.6            & 75.3            \\
\textbf{AimCLR (ours)}$^\dag$      & \textbf{83.0}  & \textbf{89.2}   & \textbf{76.4}   & \textbf{76.7}   \\ \midrule
3s-ST-GCN (AAAI $18$)               & 85.2           & 91.4            & 77.2            & 77.1            \\
3s-CrosSCLR (CVPR $21$)             & 86.2           & 92.5            & \textbf{80.5}   & 80.4            \\
\textbf{3s-AimCLR (ours)}           & \textbf{86.9}  & \textbf{92.8}   & 80.1            & \textbf{80.9}   \\ \bottomrule
\end{tabular}
\caption{Finetuned results on NTU-60 and NTU-120 dataset. ``$\dag$'' means using the bone stream data.}
\label{finetune}
\end{table}

\section{Conclusion}\label{section5}

In this paper, AimCLR is proposed to explore the novel movement patterns brought by extreme augmentations. Specifically, the extreme augmentations and the energy-based attention-guided drop module are proposed to bring novel movement patterns to improve the universality of the learned representations. The D$^3$M Loss is proposed to minimize the distribution divergence in a more gentle way. In order to alleviate the irrationality of the positive set, the nearest neighbors mining strategy is further proposed to make the learning process more reasonable. Experiments show that 3s-AimCLR significantly performs favorably against state-of-the-art methods under a variety of evaluation protocols with observed higher quality action representations.

\bibliography{aaai22}

\begin{thebibliography}{46}
\providecommand{\natexlab}[1]{#1}

\bibitem[{Cao et~al.(2019)Cao, Hidalgo, Simon, Wei, and Sheikh}]{openpose}
Cao, Z.; Hidalgo, G.; Simon, T.; Wei, S.-E.; and Sheikh, Y. 2019.
\newblock OpenPose: Realtime multi-person 2D pose estimation using part
  affinity fields.
\newblock \emph{IEEE Transactions on Pattern Analysis and Machine Intelligence
  (TPAMI)}, 43(1): 172--186.

\bibitem[{Chen et~al.(2020{\natexlab{a}})Chen, Kornblith, Norouzi, and
  Hinton}]{SimCLR}
Chen, T.; Kornblith, S.; Norouzi, M.; and Hinton, G. 2020{\natexlab{a}}.
\newblock A simple framework for contrastive learning of visual
  representations.
\newblock In \emph{International Conference on Machine Learning (ICML)},
  1597--1607.

\bibitem[{Chen et~al.(2020{\natexlab{b}})Chen, Fan, Girshick, and He}]{mocov2}
Chen, X.; Fan, H.; Girshick, R.; and He, K. 2020{\natexlab{b}}.
\newblock Improved baselines with momentum contrastive learning.
\newblock \emph{arXiv preprint arXiv:2003.04297}.

\bibitem[{Chen et~al.(2021)Chen, Li, Yang, Li, and Liu}]{MST-GCN}
Chen, Z.; Li, S.; Yang, B.; Li, Q.; and Liu, H. 2021.
\newblock Multi-scale spatial temporal graph convolutional network for
  skeleton-based action recognition.
\newblock In \emph{AAAI Conference on Artificial Intelligence}, volume~35,
  1113--1122.

\bibitem[{Cheng et~al.(2020)Cheng, Zhang, Cao, Shi, Cheng, and Lu}]{decoupling}
Cheng, K.; Zhang, Y.; Cao, C.; Shi, L.; Cheng, J.; and Lu, H. 2020.
\newblock Decoupling GCN with dropgraph module for skeleton-based action
  recognition.
\newblock In \emph{European Conference on Computer Vision (ECCV)}, 536--553.

\bibitem[{Du, Fu, and Wang(2015)}]{XYZ1}
Du, Y.; Fu, Y.; and Wang, L. 2015.
\newblock Skeleton based action recognition with convolutional neural network.
\newblock In \emph{Asian Conference on Pattern Recognition (ACPR)}, 579--583.

\bibitem[{Du, Wang, and Wang(2015)}]{HB-RNN}
Du, Y.; Wang, W.; and Wang, L. 2015.
\newblock Hierarchical recurrent neural network for skeleton based action
  recognition.
\newblock In \emph{IEEE Conference on Computer Vision and Pattern Recognition
  (CVPR)}, 1110--1118.

\bibitem[{Dwibedi et~al.(2021)Dwibedi, Aytar, Tompson, Sermanet, and
  Zisserman}]{nnclr}
Dwibedi, D.; Aytar, Y.; Tompson, J.; Sermanet, P.; and Zisserman, A. 2021.
\newblock With a little help from my friends: Nearest-neighbor contrastive
  learning of visual representations.
\newblock \emph{arXiv preprint arXiv:2104.14548}.

\bibitem[{Fang et~al.(2017)Fang, Xie, Tai, and Lu}]{alphapose}
Fang, H.-S.; Xie, S.; Tai, Y.-W.; and Lu, C. 2017.
\newblock RMPE: Regional multi-person pose estimation.
\newblock In \emph{IEEE International Conference on Computer Vision (ICCV)},
  2334--2343.

\bibitem[{Gidaris, Singh, and Komodakis(2018)}]{rotate}
Gidaris, S.; Singh, P.; and Komodakis, N. 2018.
\newblock Unsupervised representation learning by predicting image rotations.
\newblock \emph{arXiv preprint arXiv:1803.07728}.

\bibitem[{He et~al.(2020)He, Fan, Wu, Xie, and Girshick}]{moco}
He, K.; Fan, H.; Wu, Y.; Xie, S.; and Girshick, R. 2020.
\newblock Momentum contrast for unsupervised visual representation learning.
\newblock In \emph{IEEE Conference on Computer Vision and Pattern Recognition
  (CVPR)}, 9729--9738.

\bibitem[{Hu, Shen, and Sun(2018)}]{se}
Hu, J.; Shen, L.; and Sun, G. 2018.
\newblock Squeeze-and-excitation networks.
\newblock In \emph{IEEE Conference on Computer Vision and Pattern Recognition
  (CVPR)}, 7132--7141.

\bibitem[{Ke et~al.(2017)Ke, Bennamoun, An, Sohel, and Boussaid}]{XYZ2}
Ke, Q.; Bennamoun, M.; An, S.; Sohel, F.; and Boussaid, F. 2017.
\newblock A new representation of skeleton sequences for 3D action recognition.
\newblock In \emph{IEEE Conference on Computer Vision and Pattern Recognition
  (CVPR)}, 3288--3297.

\bibitem[{Lee, Kim, and Nam(2019)}]{srm}
Lee, H.; Kim, H.-E.; and Nam, H. 2019.
\newblock SRM: A style-based recalibration module for convolutional neural
  networks.
\newblock In \emph{IEEE International Conference on Computer Vision (ICCV)},
  1854--1862.

\bibitem[{Li et~al.(2021)Li, Wang, Ni, Wang, Yang, and Zhang}]{crossclr}
Li, L.; Wang, M.; Ni, B.; Wang, H.; Yang, J.; and Zhang, W. 2021.
\newblock 3D human action representation learning via cross-view consistency
  pursuit.
\newblock In \emph{IEEE Conference on Computer Vision and Pattern Recognition
  (CVPR)}, 4741--4750.

\bibitem[{Lin et~al.(2020)Lin, Song, Yang, and Liu}]{MS2L}
Lin, L.; Song, S.; Yang, W.; and Liu, J. 2020.
\newblock MS$^2$L: Multi-task self-supervised learning for skeleton based
  action recognition.
\newblock In \emph{ACM International Conference on Multimedia (ACM MM)},
  2490--2498.

\bibitem[{Liu et~al.(2019)Liu, Shahroudy, Perez, Wang, Duan, and Kot}]{ntu120}
Liu, J.; Shahroudy, A.; Perez, M.; Wang, G.; Duan, L.-Y.; and Kot, A.~C. 2019.
\newblock NTU RGB + D 120: A large-scale benchmark for 3D human activity
  understanding.
\newblock \emph{IEEE Transactions on Pattern Analysis and Machine
  Intelligence}, 42(10): 2684--2701.

\bibitem[{Liu et~al.(2020)Liu, Song, Liu, Li, and Hu}]{pkummd}
Liu, J.; Song, S.; Liu, C.; Li, Y.; and Hu, Y. 2020.
\newblock A benchmark dataset and comparison study for multi-modal human action
  analytics.
\newblock \emph{ACM Transactions on Multimedia Computing, Communications, and
  Applications}, 16(2): 1--24.

\bibitem[{Liu, Liu, and Chen(2017)}]{PR}
Liu, M.; Liu, H.; and Chen, C. 2017.
\newblock Enhanced skeleton visualization for view invariant human action
  recognition.
\newblock \emph{Pattern Recognition}, 68: 346--362.

\bibitem[{Nie, Liu, and Liu(2020)}]{sebirenet}
Nie, Q.; Liu, Z.; and Liu, Y. 2020.
\newblock Unsupervised 3D human pose representation with viewpoint and pose
  disentanglement.
\newblock In \emph{European Conference on Computer Vision (ECCV)}, 102--118.

\bibitem[{Oord, Li, and Vinyals(2018)}]{cpc}
Oord, A. v.~d.; Li, Y.; and Vinyals, O. 2018.
\newblock Representation learning with contrastive predictive coding.
\newblock \emph{arXiv preprint arXiv:1807.03748}.

\bibitem[{Pan et~al.(2021)Pan, Song, Yang, Jiang, and Liu}]{videomoco}
Pan, T.; Song, Y.; Yang, T.; Jiang, W.; and Liu, W. 2021.
\newblock VideoMoCo: Contrastive video representation learning with temporally
  adversarial examples.
\newblock In \emph{IEEE Conference on Computer Vision and Pattern Recognition
  (CVPR)}, 11205--11214.

\bibitem[{Paszke et~al.(2019)Paszke, Gross, Massa, Lerer, Bradbury, Chanan,
  Killeen, Lin, Gimelshein, Antiga et~al.}]{pytorch}
Paszke, A.; Gross, S.; Massa, F.; Lerer, A.; Bradbury, J.; Chanan, G.; Killeen,
  T.; Lin, Z.; Gimelshein, N.; Antiga, L.; et~al. 2019.
\newblock Pytorch: An imperative style, high-performance deep learning library.
\newblock \emph{Advances in Neural Information Processing Systems (NeurIPS)},
  32: 8026--8037.

\bibitem[{Pathak et~al.(2016)Pathak, Krahenbuhl, Donahue, Darrell, and
  Efros}]{context}
Pathak, D.; Krahenbuhl, P.; Donahue, J.; Darrell, T.; and Efros, A.~A. 2016.
\newblock Context encoders: Feature learning by inpainting.
\newblock In \emph{IEEE Conference on Computer Vision and Pattern Recognition
  (CVPR)}, 2536--2544.

\bibitem[{Rao et~al.(2021)Rao, Xu, Hu, Cheng, and Hu}]{AS-CAL}
Rao, H.; Xu, S.; Hu, X.; Cheng, J.; and Hu, B. 2021.
\newblock Augmented skeleton based contrastive action learning with momentum
  LSTM for unsupervised action recognition.
\newblock \emph{Information Sciences}, 569: 90--109.

\bibitem[{Shahroudy et~al.(2016)Shahroudy, Liu, Ng, and Wang}]{ntu60}
Shahroudy, A.; Liu, J.; Ng, T.-T.; and Wang, G. 2016.
\newblock NTU RGB + D: A large scale dataset for 3D human activity analysis.
\newblock In \emph{IEEE Conference on Computer Vision and Pattern Recognition
  (CVPR)}, 1010--1019.

\bibitem[{Shi et~al.(2019)Shi, Zhang, Cheng, and Lu}]{2s-AGCN}
Shi, L.; Zhang, Y.; Cheng, J.; and Lu, H. 2019.
\newblock Two-stream adaptive graph convolutional networks for skeleton-based
  action recognition.
\newblock In \emph{IEEE Conference on Computer Vision and Pattern Recognition
  (CVPR)}, 12026--12035.

\bibitem[{Si et~al.(2019)Si, Chen, Wang, Wang, and Tan}]{AGC-LSTM}
Si, C.; Chen, W.; Wang, W.; Wang, L.; and Tan, T. 2019.
\newblock An attention enhanced graph convolutional LSTM network for
  skeleton-based action recognition.
\newblock In \emph{IEEE Conference on Computer Vision and Pattern Recognition
  (CVPR)}, 1227--1236.

\bibitem[{Song et~al.(2018)Song, Lan, Xing, Zeng, and Liu}]{ST-LSTM}
Song, S.; Lan, C.; Xing, J.; Zeng, W.; and Liu, J. 2018.
\newblock Spatio-temporal attention-based LSTM networks for 3D action
  recognition and detection.
\newblock \emph{IEEE Transactions on Image Processing (TIP)}, 27(7):
  3459--3471.

\bibitem[{Su, Liu, and Shlizerman(2020)}]{PandC}
Su, K.; Liu, X.; and Shlizerman, E. 2020.
\newblock Predict \& Cluster: Unsupervised skeleton based action recognition.
\newblock In \emph{IEEE Conference on Computer Vision and Pattern Recognition
  (CVPR)}, 9631--9640.

\bibitem[{Thoker, Doughty, and Snoek(2021)}]{isc}
Thoker, F.~M.; Doughty, H.; and Snoek, C.~G. 2021.
\newblock Skeleton-contrastive 3D action representation learning.
\newblock In \emph{ACM International Conference on Multimedia (ACM MM)}.

\bibitem[{Tian et~al.(2020)Tian, Sun, Poole, Krishnan, Schmid, and
  Isola}]{Infomin}
Tian, Y.; Sun, C.; Poole, B.; Krishnan, D.; Schmid, C.; and Isola, P. 2020.
\newblock What makes for good views for contrastive learning?
\newblock \emph{arXiv preprint arXiv:2005.10243}.

\bibitem[{Van~der Maaten and Hinton(2008)}]{tsne}
Van~der Maaten, L.; and Hinton, G. 2008.
\newblock Visualizing data using t-SNE.
\newblock \emph{Journal of Machine Learning Research}, 9(11).

\bibitem[{Vemulapalli, Arrate, and Chellappa(2014)}]{Lie-Group}
Vemulapalli, R.; Arrate, F.; and Chellappa, R. 2014.
\newblock Human action recognition by representing 3D skeletons as points in a
  lie group.
\newblock In \emph{IEEE Conference on Computer Vision and Pattern Recognition
  (CVPR)}, 588--595.

\bibitem[{Vemulapalli and Chellapa(2016)}]{Rolling}
Vemulapalli, R.; and Chellapa, R. 2016.
\newblock Rolling rotations for recognizing human actions from 3D skeletal
  data.
\newblock In \emph{IEEE Conference on Computer Vision and Pattern Recognition
  (CVPR)}, 4471--4479.

\bibitem[{Wang et~al.(2012)Wang, Liu, Wu, and Yuan}]{Actionlet}
Wang, J.; Liu, Z.; Wu, Y.; and Yuan, J. 2012.
\newblock Mining actionlet ensemble for action recognition with depth cameras.
\newblock In \emph{IEEE Conference on Computer Vision and Pattern Recognition
  (CVPR)}, 1290--1297.

\bibitem[{Wang and Qi(2021)}]{clsa}
Wang, X.; and Qi, G.-J. 2021.
\newblock Contrastive learning with stronger augmentations.
\newblock \emph{arXiv preprint arXiv:2104.07713}.

\bibitem[{Woo et~al.(2018)Woo, Park, Lee, and Kweon}]{cbam}
Woo, S.; Park, J.; Lee, J.-Y.; and Kweon, I.~S. 2018.
\newblock CBAM: Convolutional block attention module.
\newblock In \emph{European Conference on Computer Vision (ECCV)}, 3--19.

\bibitem[{Xu et~al.(2020)Xu, Rao, Hu, and Hu}]{PCRP}
Xu, S.; Rao, H.; Hu, X.; and Hu, B. 2020.
\newblock Prototypical contrast and reverse prediction: Unsupervised skeleton
  based action recognition.
\newblock \emph{arXiv preprint arXiv:2011.07236}.

\bibitem[{Yan, Xiong, and Lin(2018)}]{ST-GCN}
Yan, S.; Xiong, Y.; and Lin, D. 2018.
\newblock Spatial temporal graph convolutional networks for skeleton-based
  action recognition.
\newblock In \emph{AAAI Conference on Artificial Intelligence}, volume~32.

\bibitem[{Yang et~al.(2021{\natexlab{a}})Yang, Zhang, Li, and Xie}]{simam}
Yang, L.; Zhang, R.-Y.; Li, L.; and Xie, X. 2021{\natexlab{a}}.
\newblock SimAM: A simple, parameter-free attention module for convolutional
  neural networks.
\newblock In \emph{International Conference on Machine Learning (ICML)},
  11863--11874.

\bibitem[{Yang et~al.(2021{\natexlab{b}})Yang, Liu, Lu, Er, and Kot}]{cloud}
Yang, S.; Liu, J.; Lu, S.; Er, M.~H.; and Kot, A.~C. 2021{\natexlab{b}}.
\newblock Skeleton cloud colorization for unsupervised 3D action representation
  learning.
\newblock In \emph{IEEE International Conference on Computer Vision (ICCV)}.

\bibitem[{Zhang et~al.(2019)Zhang, Lan, Xing, Zeng, Xue, and Zheng}]{VA-RNN}
Zhang, P.; Lan, C.; Xing, J.; Zeng, W.; Xue, J.; and Zheng, N. 2019.
\newblock View adaptive neural networks for high performance skeleton-based
  human action recognition.
\newblock \emph{IEEE Transactions on Pattern Analysis and Machine Intelligence
  (TPAMI)}, 41(8): 1963--1978.

\bibitem[{Zhang, Isola, and Efros(2016)}]{color}
Zhang, R.; Isola, P.; and Efros, A.~A. 2016.
\newblock Colorful image colorization.
\newblock In \emph{European Conference on Computer Vision (ECCV)}, 649--666.

\bibitem[{Zhang(2012)}]{kinect}
Zhang, Z. 2012.
\newblock Microsoft kinect sensor and its effect.
\newblock \emph{IEEE Multimedia}, 19(2): 4--10.

\bibitem[{Zheng et~al.(2018)Zheng, Wen, Liu, Long, Dai, and Gong}]{LongGAN}
Zheng, N.; Wen, J.; Liu, R.; Long, L.; Dai, J.; and Gong, Z. 2018.
\newblock Unsupervised representation learning with long-term dynamics for
  skeleton based action recognition.
\newblock In \emph{AAAI Conference on Artificial Intelligence}, volume~32.

\end{thebibliography}

\newpage
\appendix
{\noindent\Large\textbf{Supplementary Material}}
\newline

This supplementary material contains the following details: (A) Specific implementation of the data augmentations used in AimCLR. (B) Additional visualization results.

\section{Data Augmentation}\label{section1}

In contrastive learning, augmentations for positive samples bring semantic information for the encoder to learn. Just like the traditional contrastive learning framework \cite{SimCLR, moco, mocov2}, we naturally consider the ``pattern invariant'' nature of the skeleton sequence for normal augmentations: The same skeleton sequence is randomly augmented to maintain similar action patterns, and contrastive learning can be performed to extract more effective action representations. However, those carefully designed augmentations limit the encoder to further explore the novel patterns exposed by other augmentations.

Therefore, extreme augmentations are introduced to bring more novel movement patterns to learn more general feature representation. \textbf{However, we don't want to spend a lot of effort exploring data augmentations, we hope to explore a more general framework in which extreme augmentations can introduce more novel movement patterns than normal augmentations.} According to the characteristics of the skeleton sequence and several previous works \cite{AS-CAL, crossclr}, we adopt the following data augmentations.

\textbf{Normal Augmentations $\mathcal{T}$.} One spatial augmentation \textit{Shear} and one temporal augmentation \textit{Crop} are used as the normal augmentations like SkeletonCLR and CrosSCLR \cite{crossclr}.

\textbf{Extreme Augmentations $\mathcal{{\mathcal{T}}'}$.} We introduce four spatial augmentations: \textit{Shear}, \textit{Spatial Flip}, \textit{Rotate}, \textit{Axis Mask} and two temporal augmentations: \textit{Crop}, \textit{Temporal Flip} and two spatio-temporal augmentations: \textit{Gaussian Noise} and \textit{Gaussian Blur}.

\textit{1) Shear}: The shear augmentation is a linear transformation on the spatial dimension. The shape of 3D coordinates of body joints is slanted with a random angle. The transformation matrix is defined as:

\begin{equation}
A{\rm{ = }}\left[ {\begin{array}{*{20}{c}}
{\rm{1}}&{a_{12}}&{a_{13}}\\
{a_{21}}&1&{a_{23}}\\
{a_{31}}&{a_{32}}&{\rm{1}}
\end{array}} \right],
\label{shear}
\end{equation}
where $a_{12}, a_{13}, a_{21}, a_{23}, a_{31}, a_{32}$ are shear factors sampled randomly from $[-\beta, \beta]$. $\beta$ is the shear amplitude. Here we set $\beta = 0.5$. Then the sequence is multiplied by the transformation matrix $A$ on the channel dimension.

\textit{2) Crop}: For image classification tasks, crop is a very commonly used data augmentation, because it can increase the diversity while maintaining the distinction of original samples. For the temporal skeleton sequence, specifically, we symmetrically pad some frames to the sequence and then randomly crops it to the original length. The padding length is defined as $T/\gamma$, $\gamma$ is the padding ratio and here we set $\gamma = 6$.

\textit{3) Spatial Flip}: There are two facts that the symmetry of the human body structure and the robustness of actions to symmetrical exchanges. For example, throwing with the left hand and throwing with the right hand should be considered as the same class of action as ``throw''. Thus, the skeleton sequence of each frame is symmetrically transformed based on the probability of $p = 0.5$. In particular, the position of the torso in the center of the skeleton remains unchanged, while the positions of the left and right sub-skeletons are exchanged.

\textit{4) Temporal Flip}: In \cite{PCRP}, reverse prediction is proposed to learn more high-level information (e.g. movement order) that are meaningful to human perception. Hence, temporal flip is used as one of the data augmentations. Specifically, the skeleton sequence $S = {s_1, ... ,s_T}$ is reversed to $S^{'} = {s_T, ... ,s_1}$ with the probability of $p = 0.5$.

\textit{5) Rotate}: Due to the variability of the camera position in the spatial coordinate system, we introduce random rotate to the skeleton sequence, inspired by \cite{AS-CAL}, for all joint coordinate skeleton sequences, we randomly select a main rotation axis $M \in \{ X,Y,Z\}$ and choose a random rotation angle $\left[ {0, \pi /6} \right]$, and the remaining two rotation axis randomly chooses the angle $\left[ {0, \pi /180} \right]$. This is consistent with people's general perception of movement while maintaining the movement pattern: the change in the observation perspective does not affect the action itself.

\textit{6) Axis Mask}: For the 3D skeleton sequence, we hope that the sequence projected to 2D can be used as its augmented sequence. Specifically, we randomly select an axis $M \in \{ X,Y,Z\}$ and apply the zero-mask with the probability of $p = 0.5$.

\textit{7) Gaussian Noise}: To simulate the noisy positions caused by estimation or annotation, we add Gaussian noise $\mathcal{N}(0, 0.01)$ over joint coordinates of the original sequence.

\textit{8) Gaussian Blur}:  As an effective augmentation strategy to reduce the level of details and noise of images, Gaussian blur can be applied to the skeleton sequence to smooth noisy joints and decrease action details \cite{AS-CAL}. We randomly sample $\sigma  \in \left[ {{\rm{0}}{\rm{.1,2}}} \right]$ for the Gaussian kernel, which is a sliding window with length of 15. Joint coordinates of the original sequence are blurred at 50\% chance by the kernel $G( \cdot )$ below:

\begin{equation}
G(t) = \exp ( - \frac{{{t^2}}}{{2{\sigma ^2}}})\;\;t \in \left\{ { - 7, - 6, \cdots ,6,7} \right\},
\label{gb}
\end{equation}
where $t$ denotes the relative position from the center skeleton, and the length of the kernel is set to 15 corresponding to the total span of $t$.

\section{Visualization Results}\label{section2}

\textbf{Qualitative Results.} We apply t-SNE \cite{tsne} with fix settings to show the embedding distribution of SkeletonCLR and our AimCLR of PKU-MMD part I in Figure \ref{tsne}. We can clearly see that for three different streams, our AimCLR can always better make the feature representation of the same class more compact and that of different classes more distinguishable. It is also evidence of the superior performance of our AimCLR on various downstream tasks.

\begin{figure*}[tbp!]
\centering
\subfigure[SkeletonCLR]{\includegraphics[width=4cm]{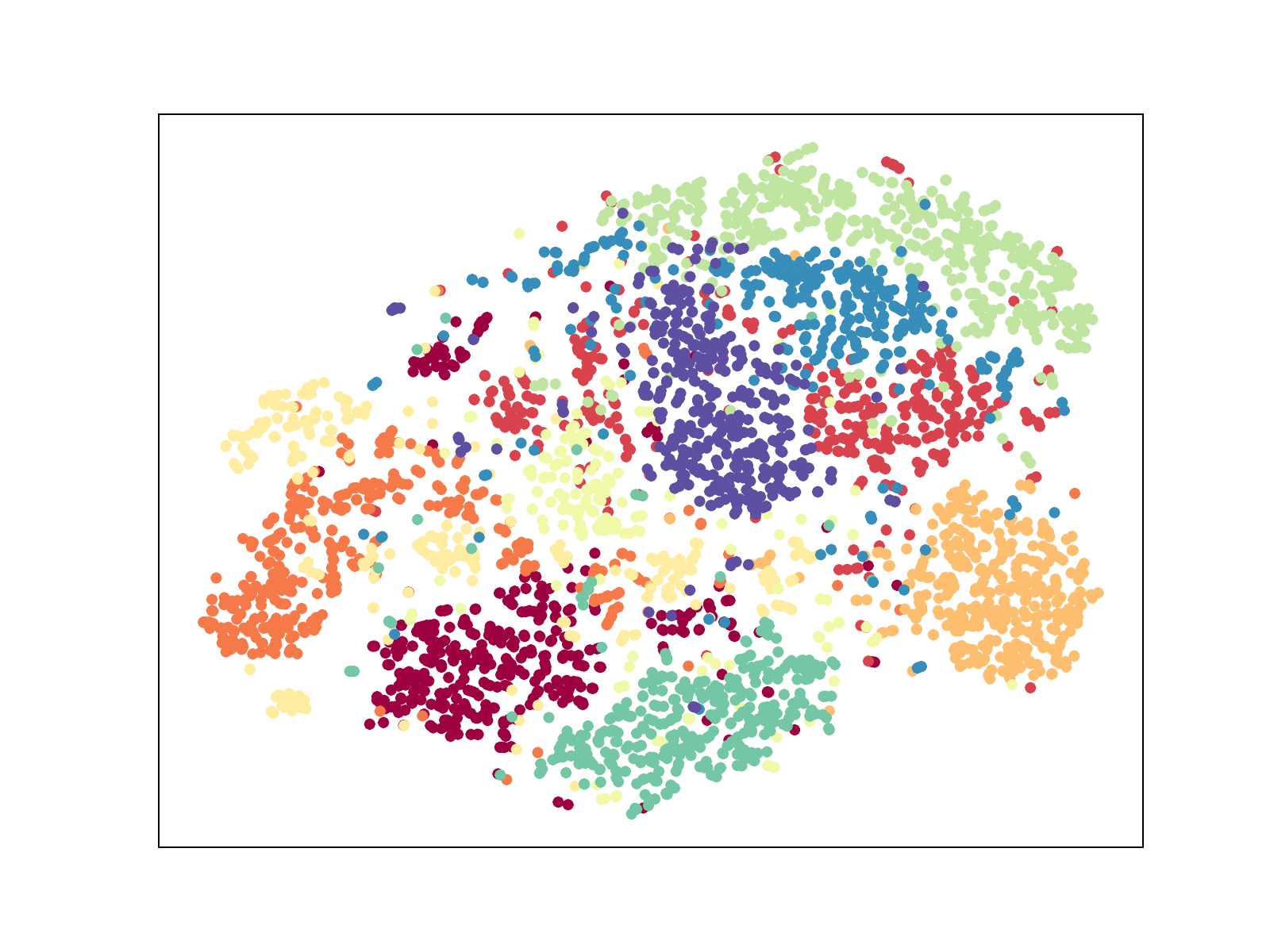}}
\subfigure[SkeletonCLR$^\dag$]{\includegraphics[width=4cm]{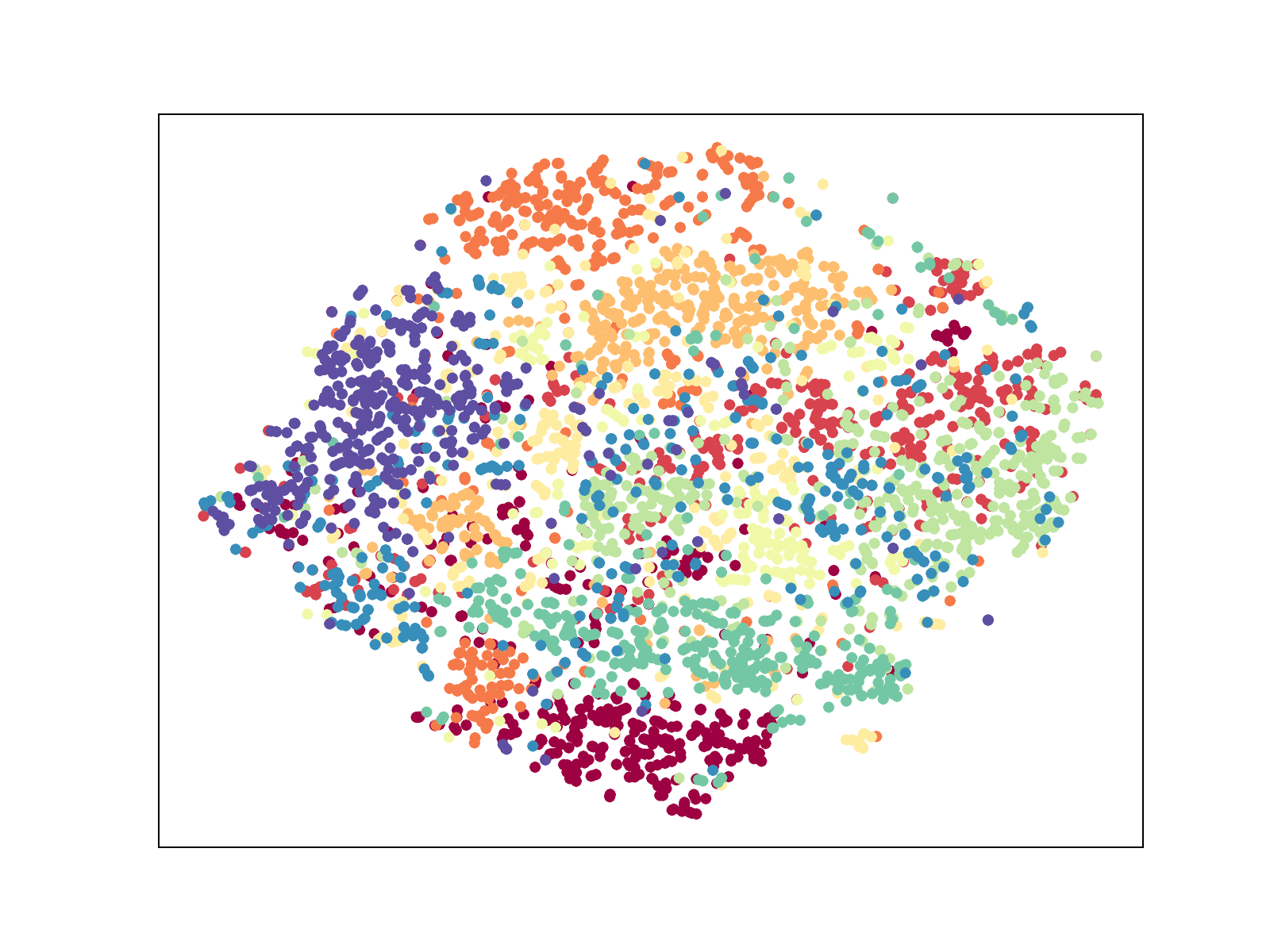}}
\subfigure[SkeletonCLR$^\ddag$]{\includegraphics[width=4cm]{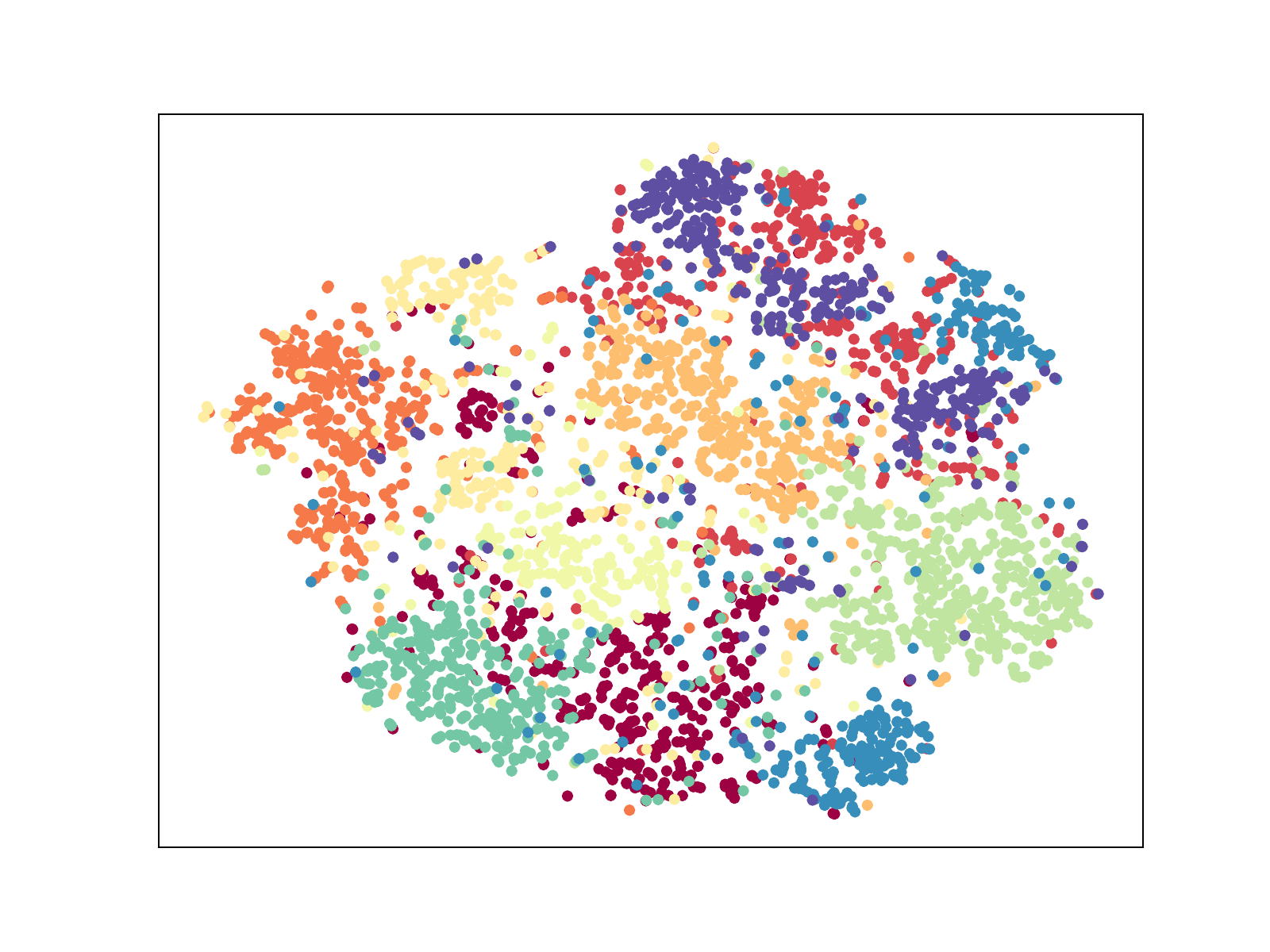}}
\subfigure[3s-SkeletonCLR]{\includegraphics[width=4cm]{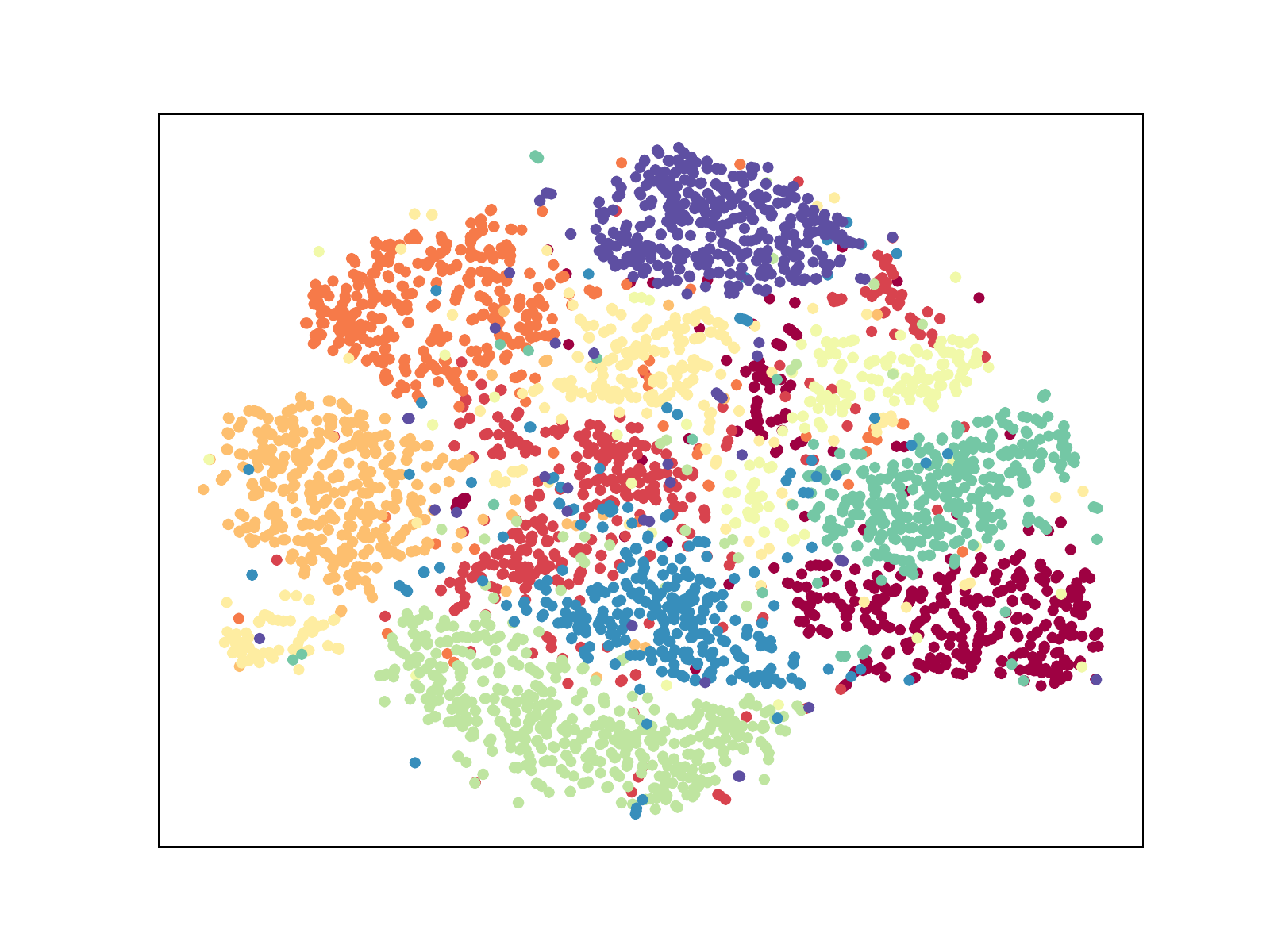}}
\\
\centering
\subfigure[AimCLR]{\includegraphics[width=4cm]{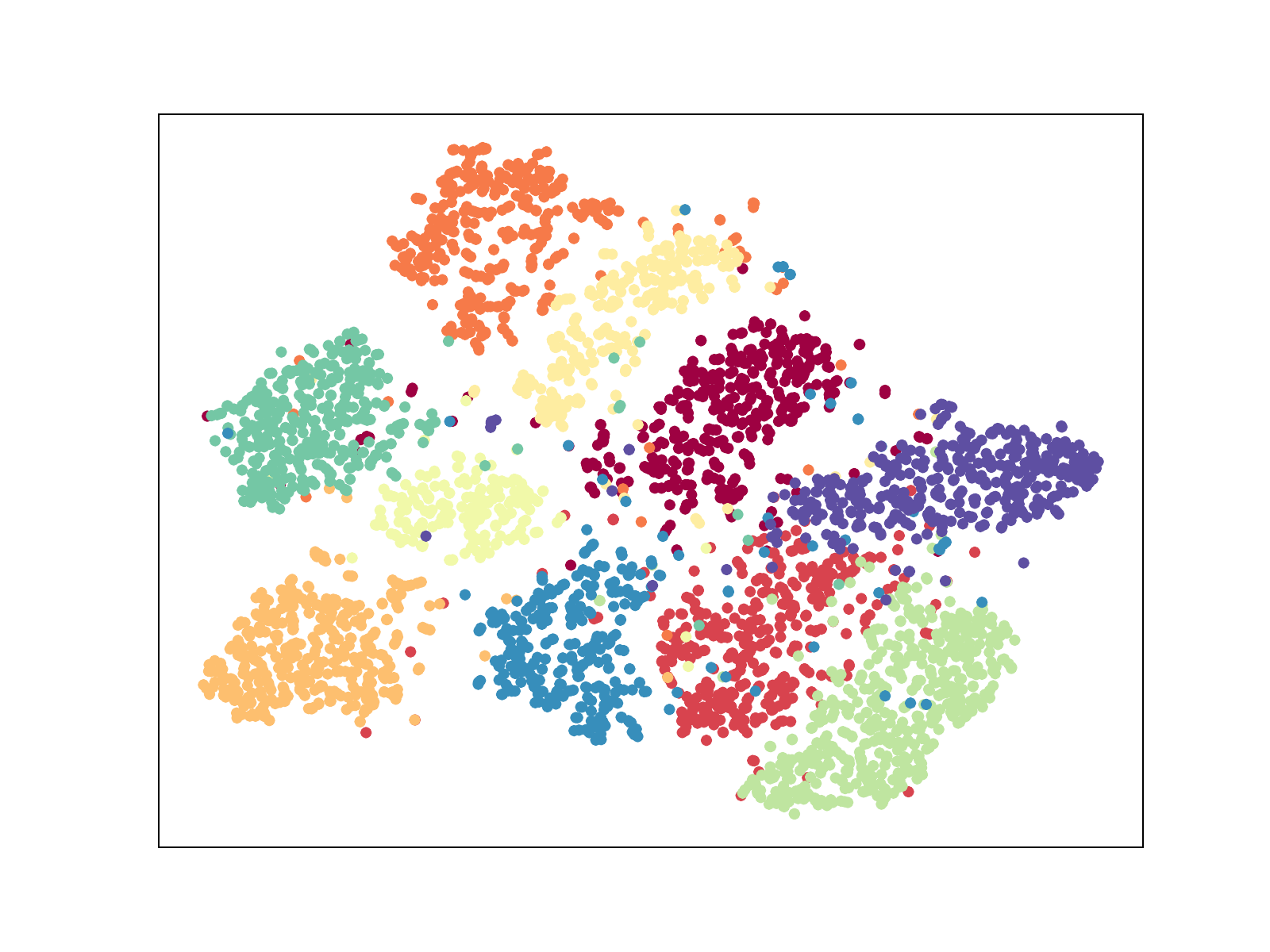}}
\subfigure[AimCLR$^\dag$]{\includegraphics[width=4cm]{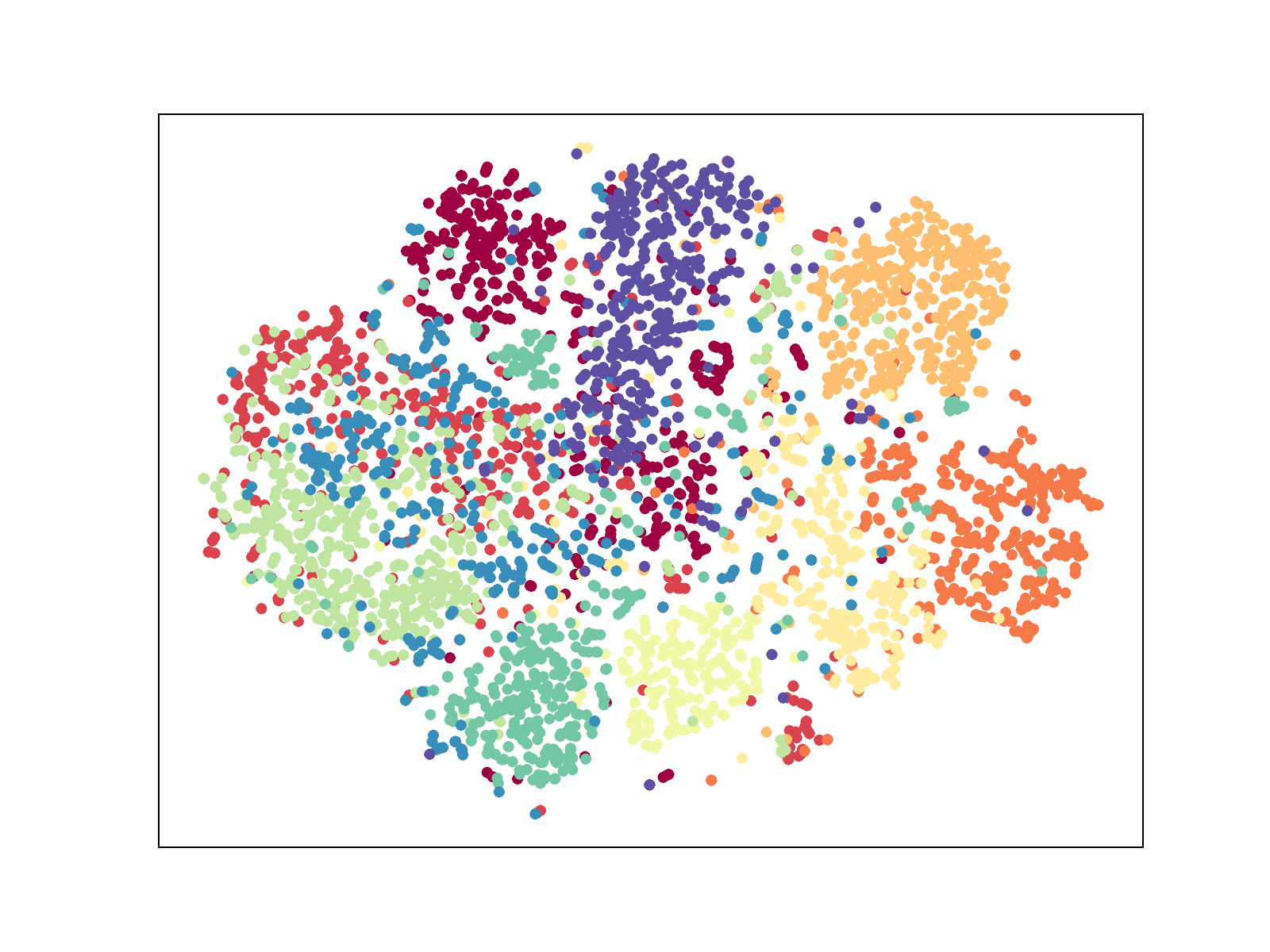}}
\subfigure[AimCLR$^\ddag$]{\includegraphics[width=4cm]{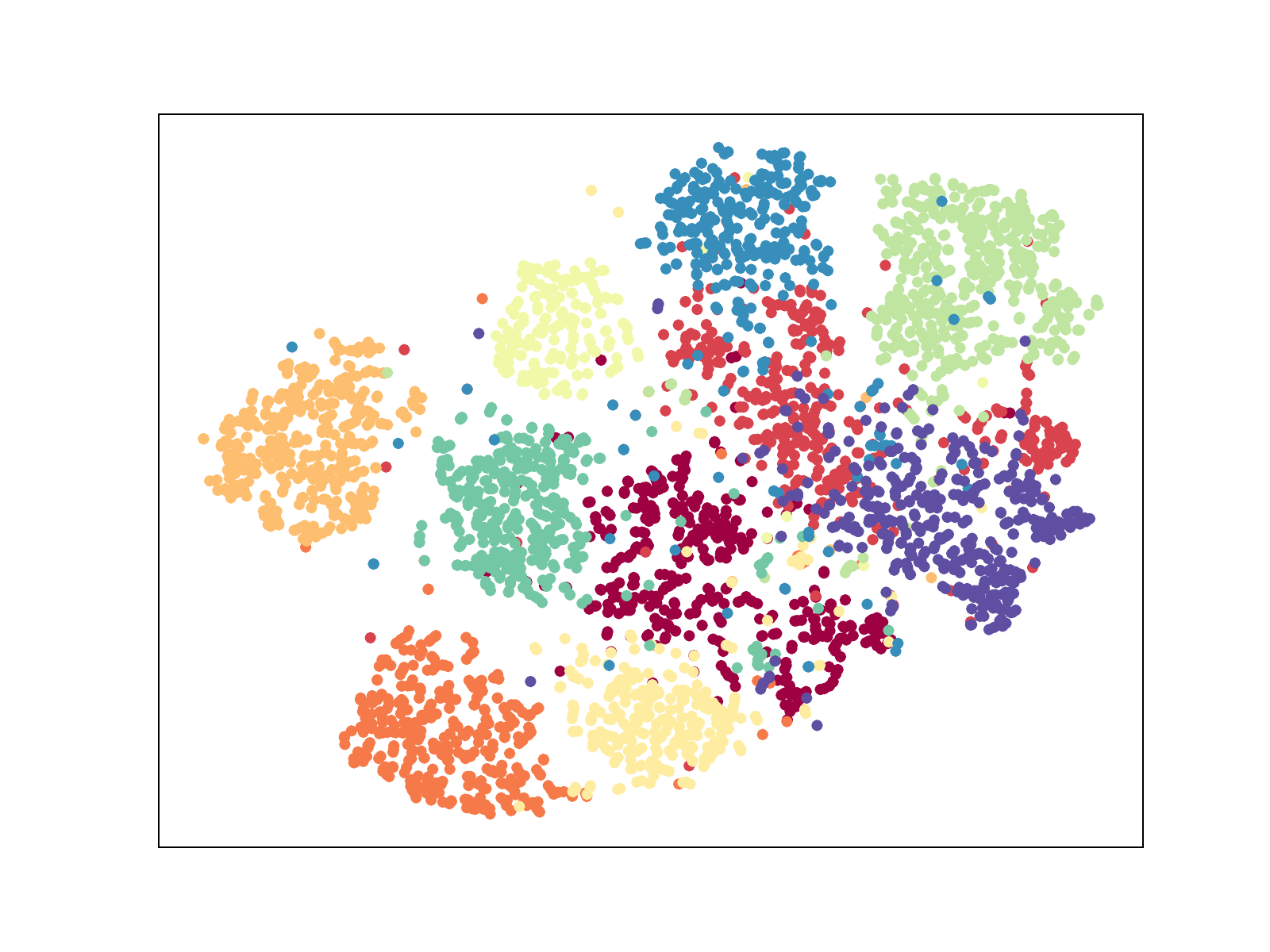}}
\subfigure[3s-AimCLR]{\includegraphics[width=4cm]{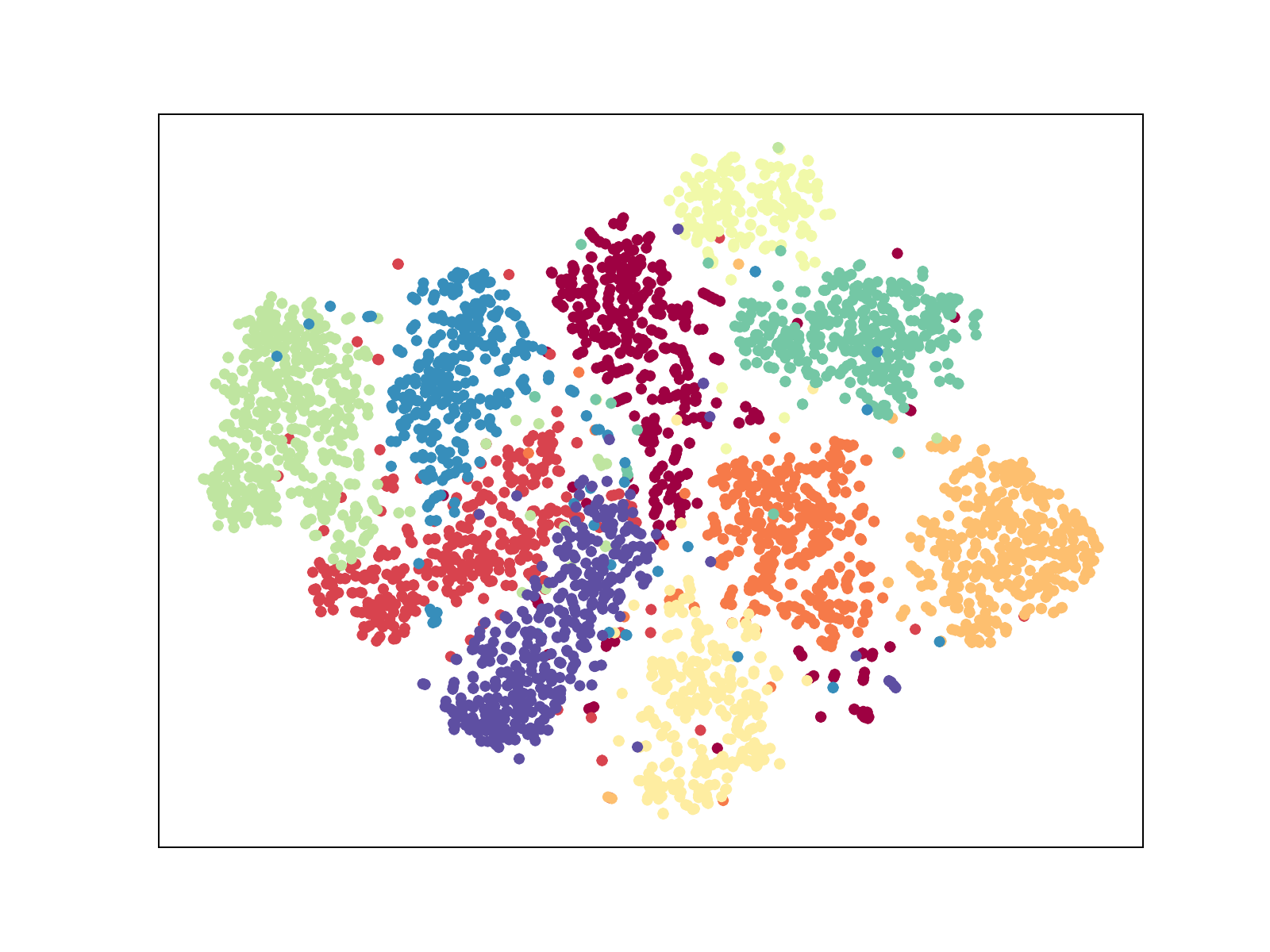}}
\caption{The t-SNE visualization of embeddings on PKU-MMD part I. ``$\dag$'' means using motion stream data while ``$\ddag$'' means using bone stream data.}
\label{tsne}
\end{figure*}

\end{document}